  \documentclass[final,5p,times,twocolumn]{elsarticle}

\usepackage{amssymb}

\usepackage{subcaption}
\usepackage{caption}
\usepackage{xurl}
\usepackage{color}
\usepackage{comment}
\usepackage{xcolor}
\usepackage{float}
\usepackage{longtable}
\usepackage{adjustbox}
\usepackage{booktabs}

\journal{.}  

\begin{document}

\begin{frontmatter}



\title{Language models for longitudinal analysis of abusive  content in Billboard Music Charts}


\author[inst1]{Rohitash Chandra}

\author[inst1]{Yathin Suresh}
\author[inst2]{Divyansh Raj Sinha}
\author[inst2]{Sanchit Jindal}

\affiliation[inst1]{organization={Transitional Artificial Intelligence Research Group, School of Mathematics and Statistics},
            addressline={UNSW Sydney}, 
            postcode={NSW 2006}, 
            country={Australia}}

\affiliation[inst2]{organization={Department of Electrical Engineering},
            addressline={Indian Institute of Technology Delhi},  
            state={Delhi},
            country={India}} 
\begin{abstract}

There is no doubt that there has been a drastic increase in abusive and sexually explicit content in music, particularly in Billboard Music Charts. However, there is a lack of studies that validate the trend for effective policy development, as such content has harmful behavioural changes in children and youths.  In this study, we utilise deep learning methods to analyse songs (lyrics) from Billboard Charts of the United States in the last seven decades. We provide a longitudinal study using deep learning and language models and review the evolution of content using sentiment analysis and abuse detection, including sexually explicit content. Our results show a significant rise in explicit content in popular music from 1990 onwards. Furthermore, we find an increasing prevalence of songs with lyrics containing profane, sexually explicit, and otherwise inappropriate language. The longitudinal analysis of the ability of language models to capture nuanced patterns in lyrical content, reflecting shifts in societal norms and language use over time.

\end{abstract}


\end{frontmatter}


\section{Introduction}
\label{sec:Introduction}


In the contemporary digital era, children's exposure to social platforms and various media forms has markedly increased, profoundly influencing their developmental trajectory \cite{livingstone2014annual}. This exposure invariably includes access to diverse musical content, and drastic changes in content question their suitability. It is imperative to analyse the lyrics of songs, given the significant role music plays in shaping emotions and personality \cite{schafer2013psychological,croom2015music} of young audiences.
Although predominantly a source of entertainment and education, music can harbour content that may adversely affect mental health \cite{becker2004deep} and influence behaviour if it contains inappropriate elements  \cite{schafer2013psychological}. Traditional content filtration methods typically rely on predefined lists of inappropriate words and phrases and basic pattern recognition techniques. Although these methods can be somewhat effective, they have several notable limitations. They struggle with evolving slang, creative language use, and contextual nuances, often resulting in ineffective detection of inappropriate content in music \cite{doi:10.1177/2053951720943234,National_Public_Radio}.
Therefore, there is a need for a robust and adaptable method for content detection so that authorities can develop policies and also advise the community on parental control and intervention.

Despite the promise of automated content filtering systems, several challenges complicate their development and deployment. One major issue is the inherent subjectivity in determining what constitutes "inappropriate" content \cite{benveniste1971subjectivity}; hence, it is difficult to create a universally acceptable filtering standard. Gudykunst et al.\ \cite{gudykunst1996influence} presented a study about the influence of culture, region, and individual differences on perceived appropriateness in communication. The study examined how various dimensions of communication, including verbal and non-verbal behaviours, can be interpreted across regional and cultural groups and individuals. This complexity necessitates more nuanced and adaptable approaches to content filtration, especially in globally distributed media such as music. 
Myers et al. \cite{myers2018differences} presented a study about the subjective nature of cultural perceptions regarding content appropriateness. They highlighted that cultural subjectivity can lead to either over-censorship, where benign content is wrongly flagged, or under-censorship, where harmful content slips through the cracks and balancing them is a critical and ongoing challenge in designing content detection algorithms. The study delves into the complexities of these cultural differences and underscores the need for nuanced approaches in content filtration systems.

Deep learning \cite{lecun2015deep} has revolutionised the field of natural language processing (NLP) \cite{nadkarni2011natural} by enabling the development of highly sophisticated models capable of understanding and generating human language with remarkable accuracy. Among these advancements, Large Language Models (LLMs) such as BERT (Bidirectional Encoder Representations from Transformers) \cite{devlin2018bert} have stood out due to their transformative impact.

Longitudinal data analysis \cite{hedeker2006longitudinal} is a powerful statistical method used to examine data collected over time, allowing researchers to track changes and identify trends within a given population. This approach is particularly valuable in fields such as social sciences, healthcare, and education, where understanding the dynamics of change over time is crucial. By analysing repeated measurements from the same subjects, longitudinal data analysis can reveal causal relationships and temporal patterns that cross-sectional studies cannot. Techniques such as mixed-effects models, growth curve modelling, and time-series analysis are commonly employed to account for the within-subject correlation and to handle missing data, which are typical challenges in longitudinal studies \cite{singer2003applied}. The individual trajectories and the impact of time-varying covariates make it indispensable for exploring developmental processes, treatment effects, and policy impacts over extended periods. Therefore, longitudinal data analysis is useful for understanding the music industry's evolution based on song lyrics.



Our study delves into a variety of research fields pertinent, including sentiment analysis, abuse detection, longitudinal analysis and ethical considerations. In abuse detection, existing studies illustrate the evolution of automated content detection from simple keyword spotting to deep learning models and LLMs. The application of these context aware models to song lyrics represent a significant advancement in the
field. However, challenges remain, particularly in ensuring these systems are unbiased and accurately reflect the diverse ways language is used in different cultural and social contexts. Although this has been attempted to be addressed by either expanding the dataset distribution or incorporating heuristics for feature extraction or label generation, these solutions are partial and incomplete.



Current studies within the space identify the temporal shift of vocabulary used in songs at a generic positive/negative level, but do not identify more nuanced emotional shifts. In parallel, the scope of longitudinal abuse studies are quite restrictive, focusing on specific genres or aspects such as alcohol use or illegal substance abuse instead of addressing broader definitions of abusive content. In the case that they do address these issues, they do not investigate current patterns and trends due to the relatively outdated nature of these studies.

Therefore, the current challenges can be summarised as follows:
\begin{enumerate}
\item{Lack of frameworks and tools that can effectively mitigate the bias in datasets.}
\item{Lack of historical and time series-based analysis to determine the temporal trends and changes in regards to lyrical content.}
\item{Lack of range and complexity in regards to feature space for sentiment analysis.}
\end{enumerate}

In this study, we utilise deep learning methods to analyse songs (lyrics) from Billboard Charts in the last seven decades. We provide a longitudinal study using deep learning and LLMs and review the evolution of content using sentiment analysis and abuse detection, including sexually explicit content.  We curate the Billboard music charts spanning the previous seven decades to serve as a dataset for longitudinal analysis and to categorise songs into explicit or non-explicit content categories. We utilise pre-trained LLMs and fine-tune them with sentiment analysis and abusive detection datasets. 
Therefore, our contribution lies in the area of automated content filtering systems that can enable parents, educators, and policy makers to make informed decisions about a safer musical environment and education system.

The structure of this study is as follows: Section 2 presents a background and literature review of related work. Section 3 presents the proposed methodology, and Section 4 presents experiments and results. Section 5 provides a discussion, and Section 6 concludes the paper with a discussion of future work.

\section{ Related Work}
\label{sec: Related Work}

\subsection{Music and human psychology}


Past studies highlighted the psychological impact of music genre and suggest that the genre and music categorisation matter more than the actual semantics (abusive) in the lyrics.
Fried  \cite{fried1999s} revealed that when participants were exposed to violent lyrical passages, they reacted more negatively if the excerpt was from a rap song instead of another genre.
Although it has been nearly two decades  Susino and Schubert \cite{susino2019cultural} had participants listen to a variety of genres, including heavy metal and hip hop, and reported that the emotions associated to each genre aligned with the broader stereotypical sentiments that are associated to the corresponding culture. They further consolidated these in a consequent study  \cite{susino2019negative} that revealed that after participants listened to two hip hop or heavy metal stimuli for a pop control stimuli, they perceived the former two as expressing more negative emotions than the pop sample. In a similar study that used the country genre instead of pop, Dunbar et al. \cite{dunbar2016threatening} validated the same findings, showcasing that participants believed that identical lyrics were more offensive when portrayed as rap. Separately, Michaela Vystrčilová and Ladislav Peška \cite{vystrvcilova2020lyrics} investigated the use of audio or lyrical features in the context of music recommender systems and found them comparable in terms of performance.  

\subsection{Music content analysis }

Manual content labelling on music platforms involves human experts (reviewers) listening to songs and reviewing lyrics to flag explicit content, such as profanity, sexual references, and violence. They follow established guidelines to categorise songs into explicit or non-explicit groups, which is then reflected in the song's metadata for user awareness \cite{ParentalAdvisoryLabel}. Despite its intention to protect younger audiences, this method suffers from subjectivity in judgement, and a lack of scalability due to the sheer volume of new music. Furthermore,  high operational costs, potential delays in content labelling, and challenges in keeping pace with constantly evolving language and slang may result in inconsistencies and oversight. Rule-based systems that involve keyword spotting techniques automate content filtering in music by using manually created rules and keyword lists to identify explicit content \cite{vygon2021learning}. However, they often misinterpret lyrics due to a lack of contextual understanding, leading to false positives and negatives. Despite being more scalable than manual methods, they require constant updates and human oversight to address new slang and cultural references, and they struggle with the nuanced language of lyrics.

Automated content detection, particularly for offensive and explicit language, has seen significant advancements in recent years. Davidson et al. \cite{davidson2017automated} explored the distinction between hate speech and offensive language, emphasising the importance of accurately differentiating between the two due to the significant legal and moral implications. Their study highlighted the limitations of lexical methods, which, while effective at identifying potentially offensive terms, often misclassify hate speech due to the nuanced context in which language is used, which underscores the need for more sophisticated methods to understand context beyond simple keyword matching   
Corazza et al.  \cite{corazza2020multilingual} proposed a modular neural network design for detecting hate speech across multiple languages, incorporating features like social media network characteristics, text-based features, word embeddings, and emotion lexicons. This approach demonstrated the value of combining different types of data to improve the accuracy of hate speech detection. Similarly, Carta et al. \cite{carta2019supervised} used a supervised multi-class multi-label word embedding approach to classify toxic comments on Wikipedia, showing that word embeddings could outperform traditional bag-of-words models in capturing the nuanced nature of abusive language.


Specific to song lyrics, Chin et al. \cite{chin2018explicit} tackled the detection of offensive content in Korean songs using a large corpus and a profanity dictionary from Namu-wiki with selected machine learning models. Chen et al. \cite{Chen2023ANovel} focused on the automated detection of explicit content in non-English languages, demonstrating the utility of combining dictionary-based methods with machine learning models. Kim and Mun \cite{kim2019hybrid}  developed a lexicon-based filtering approach using a Hierarchical Attention Network (HAN) and a Recurrent Neural Network (RNN)-based model to process Korean song lyrics and reported the importance of context in understanding explicit content which keyword-based approaches miss out.  Fell et al. \cite{fell2019comparing} conducted a comprehensive study comparing various machine and deep learning models for English music.  Overall, these studies illustrate the evolution of automated content detection from simple keyword spotting to advanced machine learning and deep learning models. The application of these models for the analysis of song lyrics, particularly for understanding context and nuance, represents a significant advancement in the field. However, challenges remain, particularly in ensuring these systems are unbiased and accurately reflect the diverse ways language is used in different cultural and social contexts.

\subsection{Abuse detection and sentiment analysis of songs}

Traditional content filtration methods typically rely on predefined lists of inappropriate words and phrases and basic pattern recognition techniques. Although these methods can be somewhat effective, they have several notable limitations. They struggle with evolving slang, creative language use, and contextual nuances, often resulting in ineffective detection of inappropriate content in songs\cite{doi:10.1177/2053951720943234,National_Public_Radio}. Rule-based systems that involve keyword spotting techniques automate content filtering in music by using manually created rules and keyword lists to identify explicit content \cite{vygon2021learning}. However, they often misinterpret lyrics due to a lack of contextual understanding, leading to false positives and negatives. Despite being more scalable than manual methods, they require constant updates and human oversight to address new slang and cultural references, and they struggle with the nuanced language of lyrics.

Automated content detection, particularly for offensive and explicit language, has seen significant advancements in recent years. Davidson et al. \cite{davidson2017automated} explored the distinction between hate speech and offensive language, emphasising the importance of accurately differentiating between the two due to the significant legal and moral implications. Their study highlighted the limitations of lexical methods, which, while effective at identifying potentially offensive phrases, often misclassifed hate speech due to the nuanced context. This  underscores the need for more sophisticated methods to understand context beyond simple keyword matching. 
Bergelid et al. \cite{bergelid2018classification} provided a thorough investigation of different permutations of available datasets that include lyrics and music metadata alongside models. 

Corazza et al. \cite{corazza2020multilingual} proposed a modular neural network design for detecting hate speech across multiple languages, incorporating features like social media network characteristics, text-based features, word embeddings, and emotion lexicons. This approach demonstrated the value of combining different types of data to improve the accuracy of hate speech detection. Similarly, Carta et al. \cite{carta2019supervised} used a supervised multi-class multi-label word embedding approach to classify toxic comments on Wikipedia, showing that word embeddings could outperform traditional bag-of-words models in capturing the nuanced nature of abusive language.


Specific to song lyrics, there have been a variety of classical machine and deep learning techniques applied to datasets that span across different languages and periods. Chin et al. \cite{chin2018explicit} tackled the detection of offensive content in Korean songs using a large corpus dating from 2010 to 2017 and a profanity dictionary from Namu-wiki with selected machine learning models.
Chen et al. \cite{Chen2023ANovel} focused on the automated detection of explicit content in non-English languages, demonstrating the utility of combining dictionary-based methods with machine learning models. 
Kim and Mun \cite{kim2019hybrid}  developed a lexicon-based filtering approach using a Hierarchical Attention Network (HAN) and a Recurrent Neural Network (RNN)-based model to process Korean song lyrics and reported the importance of context in understanding explicit content.  Fell et al. \cite{fell2019comparing} conducted a comprehensive study comparing various NLP and deep learning models for English music explicit content detection.

Fell et al. \cite{fell2019love} conducted a further study with further information, such as their structure segmentation, and the explicitness of the lyrics content.
Rospcher et al. \cite{rospocher2021} inspected the automatic detection of explicit song lyrics with a lean linear classification model that was developed on a massive dataset of 807,707 songs, 7.7\% of which were considered to be explicit.
Egivenia 2021 \cite{egivenia2021} investigated the incorporation of user-generated annotations as part of a dataset of 200 songs. Maringka \cite{maringka2024} deployed LSTM and BERT models to detect explicit content in English language music lyrics with each model having an accuracy of 88\% and 94\%, respectively. Bolla  \cite{bolla2024} explored the use of the Snorkel Framework to (reference) that acted as a heuristic to general labels for a selected dataset, which was then trained using machine learning models and maximised using the DistilBERT embeddings.


Extensive work has been done in the area of sentiment analysis for song lyrics, including a variety of feature extraction methods alongside a wide range of different model architectures.  The models have been trained using lyrical (text) and audio data 
\cite{he2020multi,rajesh2020musical,griffiths2021multi}.
There have also been studies involving a combination of the two to augment the quality of the dataset. Mihalcea et al. \cite{mihalcea2012lyrics} tested a text-based, audio-based and a combined audio-text model on the bespoke dataset of 100 songs using Ekman's 6 emotions \cite{ekman-six-emotions} as the targets. The study reported that the combined model had error rate reductions of up to 31\%. \cite{laurier2008multimodal}  with better performance when compared to the standalone audio and standalone text models. Yang et al. \cite{yang2008toward} employed a similar multimodal model, increasing the 4-class emotion classification accuracy from 46.6\% to 57.1\%.


\subsection{LLMs for music analysis}

Regarding the use of LLMs and related deep learning models, the following studies were conducted to assess their efficacy. Agrawal et. al \cite{agrawal2021transformer} explored the use of  XlNet\cite{yang2019xlnet} with the  MoodyLyrics and MER dataset, having F1 scores around 93\% on both datasets. Ravethy et al. \cite{revathy2023lyemobert} utilised a fine-tuned BERT model   on the Music4All dataset that was further enriched using the MER dataset's emotion tags of Happy, Sad, Angry, Relaxed, which are based on valence arousal in four emotion quadrants. Miyakawa \& Utsuro \cite{miyakawa2024} deployed GPT-4o for six-class emotion classification task.The dataset comprised 181 song lyrics prompted to the model with task ofsummarising the lyrics into the optimal amount of characters, which reported an accuracy of over 75\% classification rate. 
Jia \cite{jia-cnn-lstm} deployed a custom ensemble model that is composed  an LSTM whose input was taken from the output of a CNN model, which incorporated feature extraction.

Although LLMs are prominent, some studies postulate that some of the classical machine learning models perform better. Edmonds and Sedoc \cite{edmonds-sedoc-2021-multi} implemented a variety of models, such as a fine-tuned BERT, Naive-Bayes, and Random Forests  and reported that models trained on relatively small song datasets achieve marginally better performance than BERT finetuned on large social media or dialog datasets.


Despite this, it is irrefutable that LLMs provide great flexibility with respect to other machine learning methods, enabling the enrichment of these datasets with user annotations and guidance as well.  Choi et. al. \cite{choi2014} investigated the use of lyrics and also users' interpretation  as inputs into two BERT models  ensembled together, addressing a potential weakness of  using the lyrics alone as it provides even more context surrounding the tokens. Donelly and Beery \cite{donnelly2022evaluating} used social media discourse as a proxy for song lyrics data to determine the efficacy of using such a dataset for this learning task and achieving a moderate Pearson’s correlation for valence and arousal respectively, using BERT-based models.
Hasan et al. \cite{Hasan2014UsingHA}, on the contrary, found the usage of crowd/user-sourced information a hindrance  in regards to the labels associated  for sentiment analysis based on Twitter(X) data.

\subsection{N-gram analysis}


N-grams are widely employed in NLP for analysing text data, particularly to give a better undersining of the decision-making process for tasks such as sentiment analysis and text classification \cite{yannakoudakis1990n}. They efficiently capture sequences of words, providing insights into local dependencies within the text and enabling straightforward feature extraction. Despite their simplicity and interpretability, N-grams are limited by their inability to grasp longer context dependencies and subtle semantic nuances across texts. This hampers their applicability in understanding complex language structures and diverse domains \cite{culy2003limits}. Challenges also include managing sparse data issues with infrequent N-grams and the computational overhead of processing higher-order sequences. However, N-grams remain fundamental in NLP research and applications, as they serve as foundational models and benchmarks for evaluating more advanced techniques. They offer scalability and transparency, making them essential starting points for developing sophisticated language models and deep learning architectures, and have been effectively utilised with sentiment analysis \cite{chandra2024large,chandra2025longitudinal}.

\section{Methodology}
\subsection{Data}

We obtained a dataset for the weekly Billboard music charts from 1958 onwards from Kaggle \cite{godefroy_lambert_2024}, which comprised of about 342,000 entries. The Billboard music charts feature a list of the top 100 weekly songs, where many songs appear multiple times as they could remain in the top 100 for several weeks. We needed to eliminate repetitions for our analysis and hence, we extracted unique songs, which resulted in a dataset containing around 31,000 unique entries. We then utilised the Spotify API (Application Programmer Interface)  \cite{spotifywebapi} to assign labels indicating whether a song was explicit and the Genius API \cite{geniusapi} to extract the lyrics of these songs. During this process, we found that there was a significant disparity in the distribution of explicit labels over the years. 

\begin{figure}[h!]
    \centering
    \includegraphics[width=0.85\linewidth]{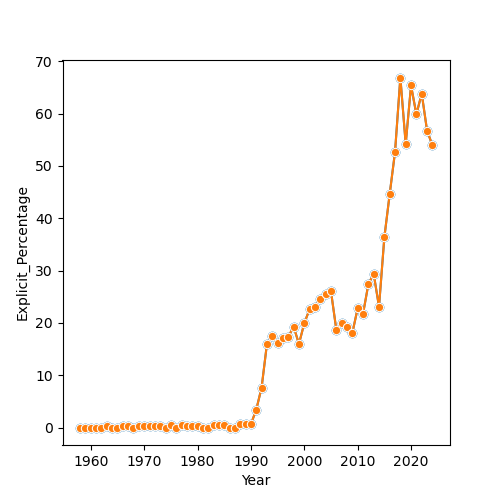}
    \caption{Percentage of Explicit Songs by Year}
    \label{fig:percentage-explicit-songs}
\end{figure}

As illustrated in Figure~\ref{fig:percentage-explicit-songs}, the percentage of explicit songs was significantly lower in the earlier years, which presented a challenge for creating a uniform training dataset. We excluded the initial years with sparse data on explicit songs from the analysis to ensure a uniform training dataset. Specifically, we removed songs released before 1990 from the dataset based on the observation that explicit labels became more prevalent in the music industry post-1990. Therefore, our refined dataset provides a more accurate reflection of trends and patterns related to explicit content in popular music.

\subsection{Data processing}

We processed the data   taking into consideration that songs could chart over multiple years, and therefore duplicates needed to be dropped after partitioning based on the date. This resulted in the dataset of 10,600 songs in total and the next step was to then extract features such as the lyrics, label and genre of the songs. Each one of these features had challenges in regard to data quality and needed to be addressed accordingly to ensure the validity of the dataset for consequent model training.

 Spotify and Genius API encountered problems by  not being able to identify songs accurately due to the formatting of artists and songs especially when multiple artists are involved in multiple songs. In regards to songs, some of them would have alternate titles or be created specifically for a movie, play or event. Examples of this include "Never Too Far/Hero Medley" by Mariah Carey or "The Cup Of Life (The Official Song Of The World Cup; France '98)" by Ricky Martin. These titles would be simply processed by removing anything within the brackets or any characters beyond and including the slash. This simple approach was less effective for artists - this was seen when they had descriptions such as "Anita Cochran (Duet With Steve Wariner)" Excluding information concerning other artists involved caused issues when it came to querying the API. In addition to this, there were a variety of tags that signified the collaboration for other artists, which caused issues with the API, such as "G-Eazy x Bebe Rexha" or "Chris Brown Feat. Ludacris". We identified these varying conventions  with regex and consequently standardised by replacing them with "Featuring" instead. 


The most difficult feature to generate was the genre of the songs which was essential to assess and derive insights regarding the model performance. This was because there were a fair few songs that simply did not have a genre associated with them through Spotify. Therefore, we used external music streaming platforms such as Deezer and Last FM via their API to ascertain the genre of songs in the case that Spotify did not contain it. Each API had their drawbacks; for Deezer the songs did not have genre classifications but rather the album itself. 
Similarly, Last FM did not have an explicit genre metadata associated to a song, but rather community created "tags". The top tag for any given song would often be the corresponding genre, but naturally, there were exceptions to the rule. As a byproduct of this, the API's were subsequently queried in the order of their reliability; Spotify followed by Deezer and finally Last FM. Despite going through this extensive process,  some songs did not have a genre associated with them as there were no records in any of the three vendors. We carried out further processing through the use of ChatGPT, which not only identified the genre of the songs with the missing values but also assisted in mapping sub-genres such as "electro" and "house" into "electronic". This process of mapping also involved the standardisation of genre names, such as converting "hip-hop" and "hip hop" into the "rap" category.

The lyrics data for the songs, which was derived from the Genius API, had a variety of phrases to signify the structure of the song alongside random promotional content regarding the purchase of artist tickets for their shows. This included sequences of sub-strings such as "[Chorus: <artist>]" or ""See Taylor Swift LiveGet tickets as low as \$69You might also like" within the lyrics itself. Regex patterns were used to remove these phrases efficiently. Once these lyrics were sourced, the next phase of data quality assurance was undertaken on the lyrics.

Due to the fact that the Billboard Top 100 charts are language agnostic, there were a significant number of entries that were non-English. Prominent genres and countries frequently in the data included "K-pop" (Korean pop) songs from Korean artists and Latin music by prolific and popular artists such as BTS and Bad Bunny from each country, respectively. We identified  non-English songs using a combination of Python's LangDetect library \footnote{\url{}} and removed them. However, to account for the restriction on the function to process only the first 5000 characters, the lyrics were chunked and parsed individually. In the case that a given chunk was identified to have a non-English language, the song was removed from the dataset.



\label{sec: Methodology}

\subsection{Deep learning-based language models}
The Long Short-Term Memory (LSTM) network is an RNN \cite{LSTM} designed to overcome the problem of long-term dependency on data in conventional RNNs. LSTM networks excel in learning and retaining information over extensive sequences, leveraging memory cells and gated mechanisms—such as forget, input, and output gates—to selectively store and access information across time steps.


BERT is an LLM with innovative architecture that processes text in both directions, allowing it to grasp the context of words more comprehensively than conventional unidirectional language models \cite{}. This bidirectional capability has significantly enhanced its performance across various NLP tasks, from sentiment analysis to text content classification. RoBERTa (Robustly optimised BERT) \cite{liu2019roberta} was developed to improve the model's performance further, which employs the same underlying architecture and incorporates several key optimisations, such as longer training time, larger batch size, and larger data corpus. These enhancements enable RoBERTa to achieve even greater accuracy and robustness in understanding complex language patterns. BERT and RoBERTa can be fine-tuned with domain-specific datasets for sentiment analysis \cite{chandra2024large} and detecting abusive language (Hate-BERT) \cite{caselli2020hatebert}. 
\begin{figure*}[t]
    \centering
    \includegraphics[width=1\linewidth]{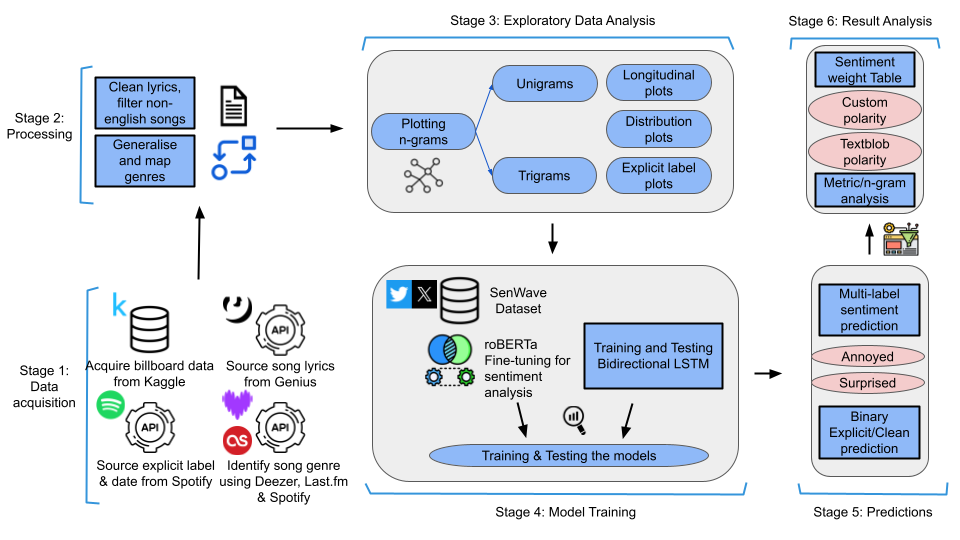}
    \caption{Framework for review of abusive and inappropriate content in Billboard Charts}
    \label{fig:framework}
\end{figure*}


\subsection{Framework for longitudinal  sentiment and abuse analysis for Billboard Charts}

Sentiment analysis datasets generally provide sentiment polarity analysis, which is not ideal for our application since we need information about the type (anger, sadness, happiness) of sentiments present in lyrics. 
The  SenWave dataset \cite{yang2020senwave} for sentiment analysis comprises
over 105 million collected tweets worldwide, assessing global sentiment fluctuations during the COVID-19 pandemic. The dataset captured tweets from a variety of languages as well including English, Spanish, French, Arabic, Italian and Chinese, from March 1st 2020, onwards. However, only a fraction of the tweets formally annotated that included 10,000 English and 10,000 Arabic tweets across ten distinct categories: optimism, gratitude, compassion, pessimism, anxiety, sadness, anger, denial, official reports, and joking. We utilise English tweets for fine-tuning the BERT-based sentiment analysis model for multi-label sentiment analysis, where one tweet can have multiple classification labels (sentiments).

 HateBERT \cite{caselli2020hatebert} is a fine-tuned variation of the BERT base-uncased model for abuse detection in text data, fine-tuned using RAL-E  (Reddit Abusive Language English), a collection of Reddit comments in English from communities banned for being either offensive, abusive, or hateful. 
At a broad level, due to the model's domain specification in abuse detection, HateBERT is suitable for our classification task of explicit song identification.
 
We present our framework for longitudinal sentiment and abuse analysis for Billboard Charts (Figure \ref{fig:framework}) that demonstrates how we efficiently collect, process and classify songs for review of sentiments, abusive and inappropriate content over the last three decades. 

In Stage 1, we extract the data for the Billboard Music charts from Kaggle, which includes information regarding all songs that have historically been charted. We then append this data using lyrics and explicit/clean labels pulled from the Genius and Spotify API, respectively. In addition to this, we compile the genre of songs from a variety of sources, including music services such as Deezer and Last.FM due to the sparsity of the genre tag within the Spotify API. 

In Stage 2, we apply several preprocessing methods to prepare data for analysis. We first filter out non-English songs using language detection techniques (specifically the detect function in the \textit{langdetect} Python library \footnote{\url{https://pypi.org/project/langdetect/}}) to focus exclusively on English-language songs, maintaining linguistic consistency. Once this is
done, we generate three distinct views on the data based on differing time scopes, with duplicates dropped based on the temporal window. Particularly, these scopes are time-agnostic, decade-based based and year-based. We then standardised the data fields, such as song titles and explicit labels, to ensure consistency across the dataset. This step also involves converting text to lowercase, removing special characters, and standardising date formats. Separately, the genre field was processed further by mapping sub-genres into the major categories, such as hip-hop and country.

Stage 3 features exploratory data analysis - this includes preliminary temporal distribution analysis with respect to the features, including genre, explicit tag and lyrics. For the latter, n-grams and an abuse look-up dictionary will be employed to analyse the data. This analysis aims to validate the high-level patterns present in the data by corroborating it with previous studies and also to confirm the mitigation of potential biases in the dataset.


Stage 4 involves the deployment of the chosen models, which includes fine-tuning BERT-based architectures on the processed Spotify/Billboard dataset from Stage 2 and the SenWaVe dataset for abuse detection and sentiment analysis, respectively.



Stage 5 involves the use of metrics such as precision, recall, and accuracy to assess model performance directly against the relevant datasets. For sentiment analysis, the model will first be analysed with respect to the fine-tuning dataset to assess if the transfer-learning was successful (i.e. the SenWaVe data). Once this has been validated, the predicted sentiments will be compressed into a polarity score and converted into a binary label based on a threshold to facilitate direct comparison with Spotify’s explicit label. For abuse detection, the abuse label is compared against Spotify’s rating directly without additional processing.

Stage 6 explores the output of the validated models with specific case studies in each genre. Furthermore, error analysis will be conducted to ascertain if the models faltered in a specific genre or era. Finally, longitudinal analysis will be applied to the
outputs of both models to understand and identify temporal trends and patterns with the progression of music content. 




\section{Experiments and Results}

\subsection{Data Analysis}

We performed exploratory data analysis to understand the high-level trends present in the data, specifically for Spotify's explicit content with an externally sourced profanity based on dictionary check.

\begin{figure}[htbp!]
    \centering
    \includegraphics[width=0.85\linewidth]{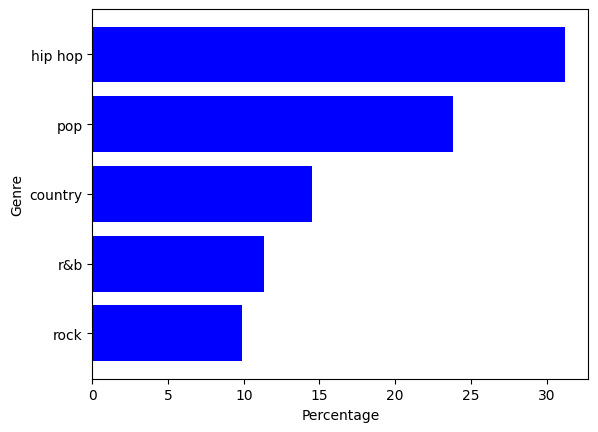}
    \caption{Distribution of songs by genre in the dataset.}
    \label{fig:genre-distribution}
\end{figure}

Figure \ref{fig:genre-distribution} shows the distribution of the songs by their genres in the dataset. We observe that the genres of"Hip Hop" and "Pop" are the largest, which reflects the shifting trend of the music industry to favour the two genres. At a broad level, the dataset is relatively balanced across all the primary genres, with hip hop, pop, country, R\&B and rock accounting for over 90\% \% of songs within the dataset. The other 10\% is attributed to fringe genres such as Gospel, Jazz and Blues, which fall in neither one of the five primary categories as described earlier.
Over the years, there has been a significantly higher amount of abusive/explicitly labelled songs for the Billboard Hot 100 charts (Figure \ref{fig:percentage-explicit-songs}). We can observe that up until 1990, only 2-3\% of songs with explicit lyrics reached the Hot 100 list. However, past this point the proportion of these explicit songs grew rapidly, reaching a maximum of 65\% in the late 2010s.

\begin{figure}[htbp!]
    \centering
    \begin{subfigure}{1\linewidth}
        \centering
        \includegraphics[width=1\linewidth]{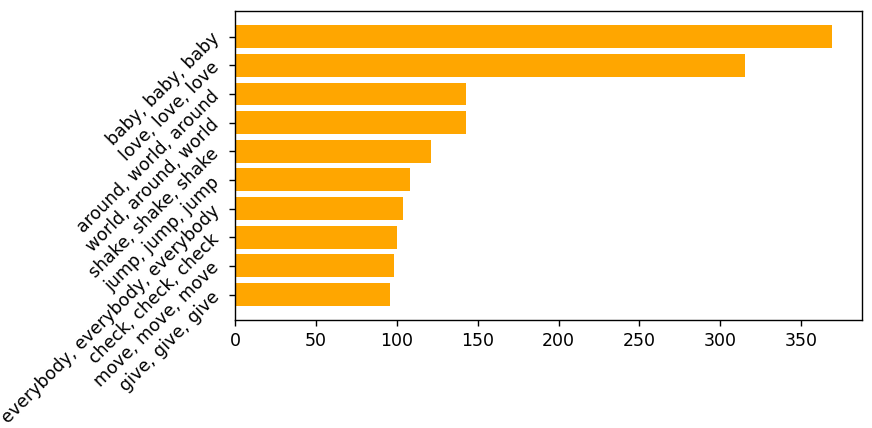}
        \caption{1990-2005}
        \label{fig:subfig1}
    \end{subfigure}
    
    \begin{subfigure}{1\linewidth}
        \centering
        \includegraphics[width=0.9\linewidth]{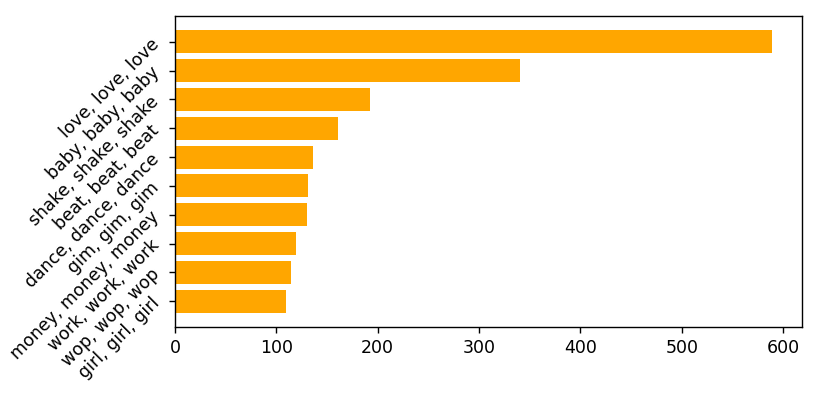}
        \caption{2006-2016}
        \label{fig:subfig2}
    \end{subfigure}
    
    \begin{subfigure}{1\linewidth}
        \centering
        \includegraphics[width=0.9\linewidth]{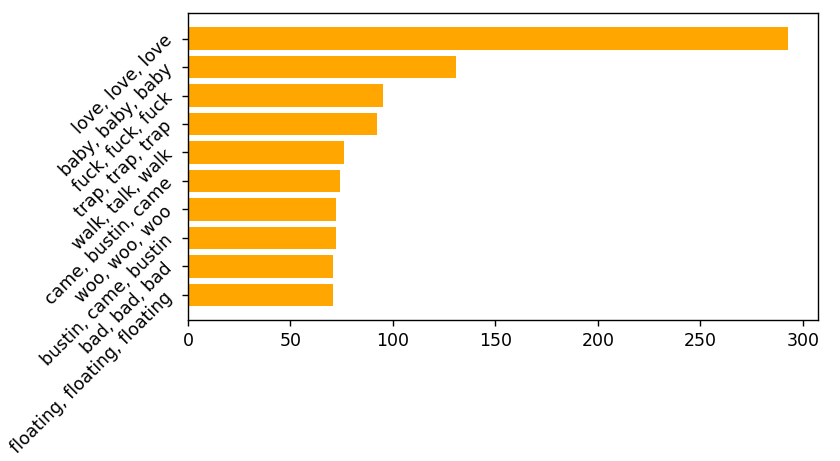}
        \caption{2017-2024}
        \label{fig:subfig3}
    \end{subfigure}
    
    \caption{Top 10 trigrams over the years: 1990 - 2024}
    \label{fig:trigrams1990-2024}
\end{figure}

We performed a longitudinal analysis of abusive and inappropriate content in Billboard Music Charts from 1990 to 2024. We present trigram analysis (Figure   \ref{fig:trigrams1990-2024}) to identify major themes and changes in lyrical content over time, consisting of three periods: 1990-2005, 2006-2016, and 2017-2024. 
We notice that in the early period (1990-2005), the dominant trigrams highlight repetitive and rhythmic elements characteristic of pop and hip-hop music. Frequent trigrams such as "baby, baby, baby," "love, love, love," and "around, world, around" indicate a strong emphasis on themes of love and catchy, repetitive phrases. Moving into the middle period (2006-2016), the focus on love and dance persists with trigrams like "love, love, love" and "dance, dance, dance." Notably, "money, money, money" emerges, indicating a growing fascination with wealth and materialism in song lyrics. There is a significant shift towards more explicit content in the latest period (2017-2024). Trigrams such as "f*ck, f*ck, f*ck" and "trap, trap, trap" become prominent, reflecting the increased use of explicit language and references to drug culture in modern music. Nevertheless, themes of love remain present, as shown by the continued presence of "love, love, love."

\subsection{Model performance}

Tables \ref{tab:model_performance} and \ref{tab:model_performance_extended} present a comparsion of  BERT and RoBERTa-based models for sentiment classification using the SenWave dataset.

\begin{figure}[htbp!]
    \centering
    \includegraphics[width=1\linewidth]{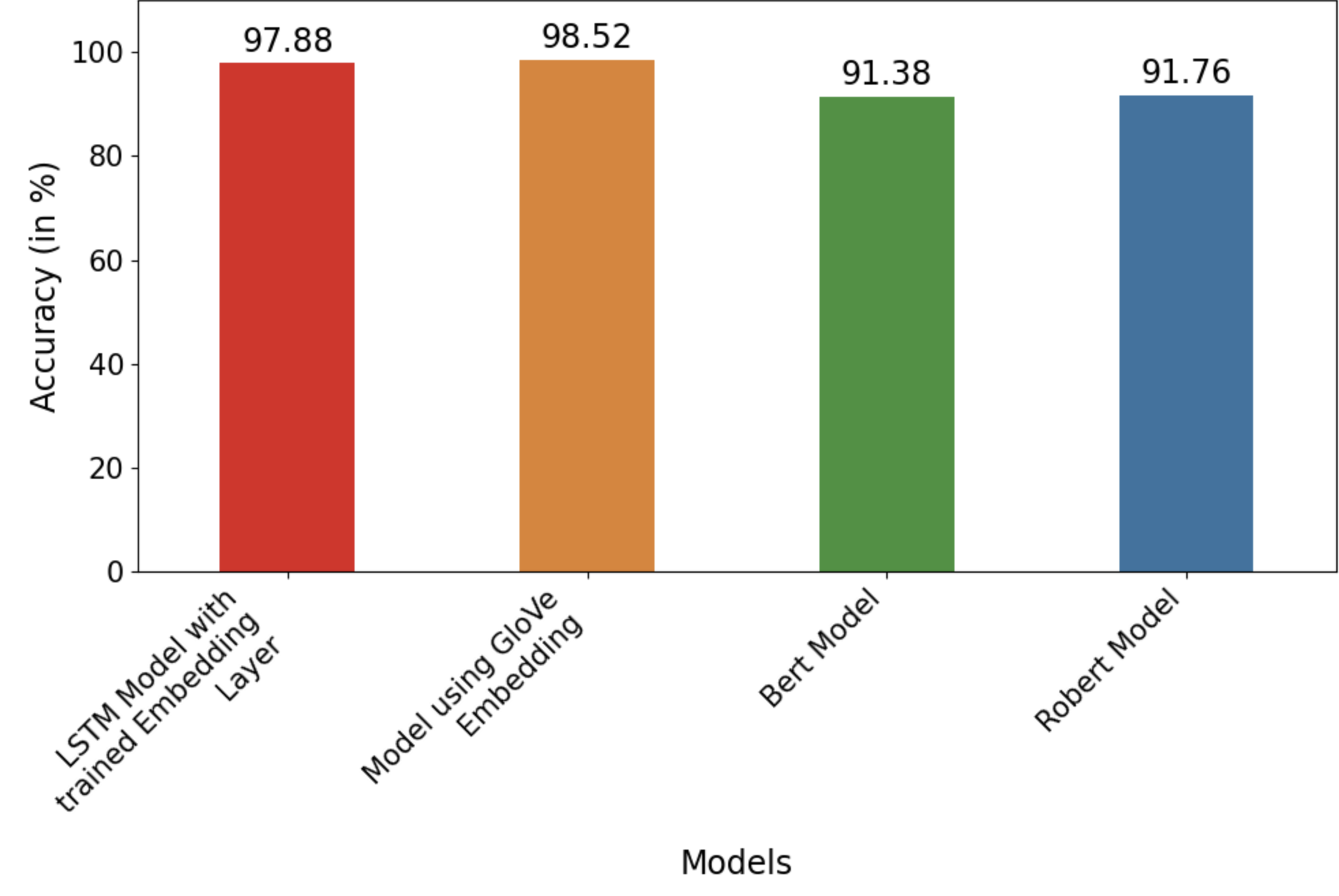}
    \caption{Comparison of model accuracy}
    \label{fig:model-accuracies}
\end{figure}

Our analysis also focuses on assessing the effectiveness of various deep learning models, each utilising different embeddings such as GloVe  \cite{pennington2014glove}, BERT, and custom-trained embeddings (see Figure \ref{fig:model-accuracies}). Our findings indicate that the model utilising GloVe embeddings achieved the highest accuracy in this classification task with an accuracy of 98.52\%. 
The performance of other models using BERT and custom-trained embeddings also gave robust classification results with accuracies of 91.4\% and 97.5\%, respectively. The variation in accuracy among different model emphasises the importance of embedding selection in NLP applications.

\begin{table*}[htbp!]
\centering
\caption{Overall Model Performance - Billboard Dataset}
\label{tab:model_performance}
\begin{tabular}{lccc}
\toprule
\textbf{Validation} & \textbf{Accuracy} & \textbf{F1 Micro} & \textbf{F1 Macro} \\
\midrule
roBERTa & 0.9135 $\pm$ 0.0019 & 0.9149 $\pm$ 0.0013 & 0.9004 $\pm$ 0.0040 \\
BERT    & 0.9119 $\pm$ 0.0024 & 0.9121 $\pm$ 0.0031 & 0.9001 $\pm$ 0.0041 \\
\midrule
\textbf{Test} & \textbf{Accuracy} & \textbf{F1 Micro} & \textbf{F1 Macro} \\
\midrule
roBERTa & 0.9176 $\pm$ 0.0023 & 0.9172 $\pm$ 0.0019 & 0.9040 $\pm$ 0.0031 \\
BERT    & 0.9138 $\pm$ 0.0025 & 0.9138 $\pm$ 0.0021 & 0.9036 $\pm$ 0.0035 \\
\bottomrule
\end{tabular}%
\end{table*}

\begin{table*}[htbp!]
\centering
\caption{Overall Model Performance - SenWaVe Dataset}
\label{tab:model_performance_extended}
\begin{tabular}{lccccc}
\toprule
\textbf{Validation} & \textbf{Hamming Loss} & \textbf{Jaccard Score} & \textbf{Label Rank Avg. Prec. Score} & \textbf{F1 Macro Score} & \textbf{F1 Micro Score} \\
\midrule
roBERTa & 0.1294 $\pm$ 0.00012 & 0.5339 $\pm$ 0.0019 & 0.7874 $\pm$ 0.0017 & 0.5416 $\pm$ 0.0035 & 0.6045 $\pm$ 0.0018 \\
BERT    & 0.1305 $\pm$ 0.0006  & 0.5413 $\pm$ 0.0015 & 0.7594 $\pm$ 0.0019 & 0.5221 $\pm$ 0.0038 & 0.6102 $\pm$ 0.0016 \\
\midrule
\textbf{Test} & \textbf{Hamming Loss} & \textbf{Jaccard Score} & \textbf{Label Rank Avg. Prec. Score} & \textbf{F1 Macro Score} & \textbf{F1 Micro Score} \\
\midrule
roBERTa & 0.1369 $\pm$ 0.0009 & 0.5144 $\pm$ 0.0025 & 0.7763 $\pm$ 0.0011 & 0.5326 $\pm$ 0.0029 & 0.5914 $\pm$ 0.0024 \\
BERT    & 0.1409 $\pm$ 0.0013 & 0.5212 $\pm$ 0.0016 & 0.7675 $\pm$ 0.0014 & 0.5234 $\pm$ 0.0031 & 0.5991 $\pm$ 0.0022 \\
\bottomrule
\end{tabular}
\end{table*}

\subsection{Sentiment Analysis}

\begin{figure}[h!]
    \centering
    \includegraphics[width=\linewidth]{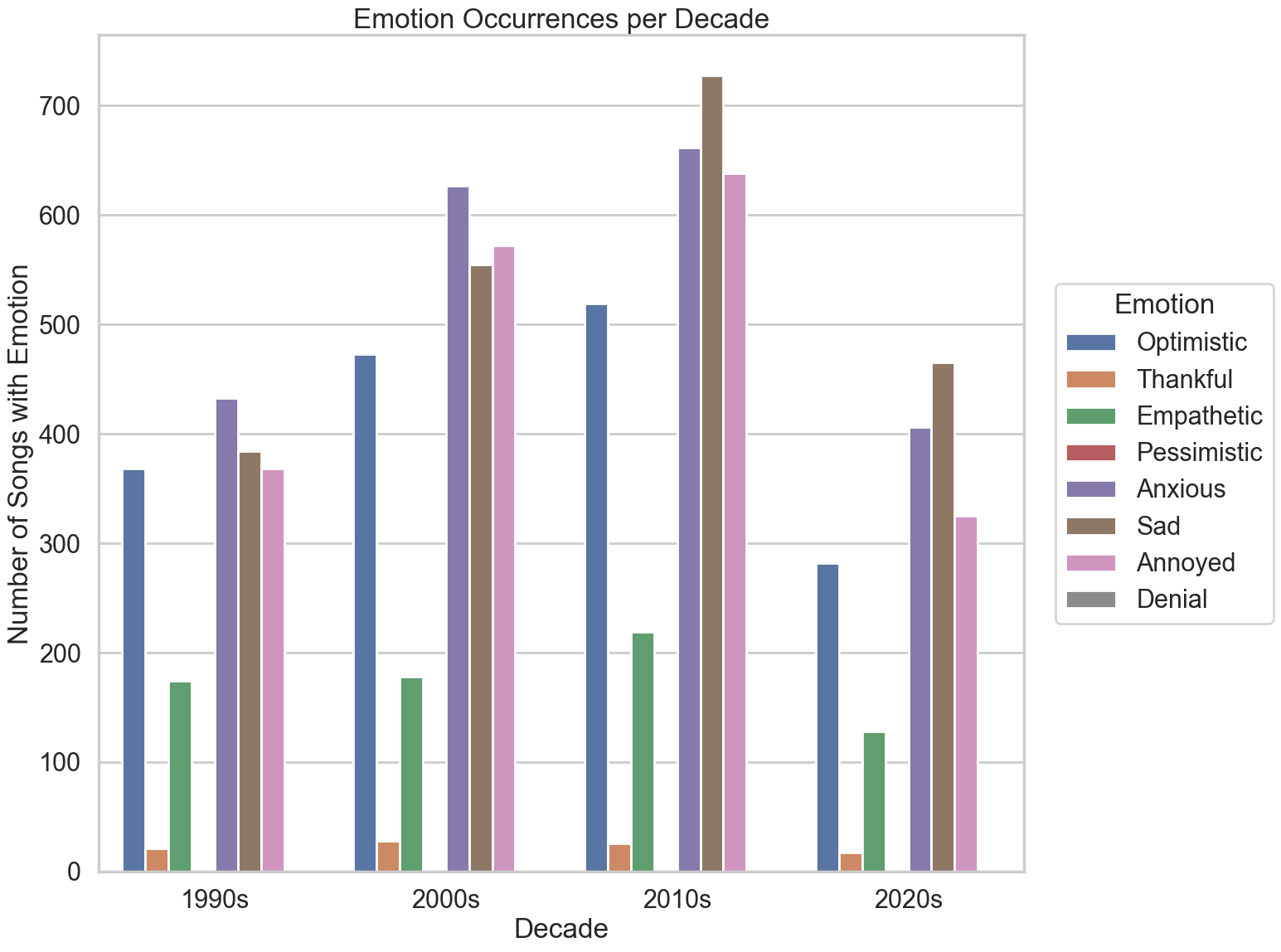}
    \caption{Emotion counts over the decades.}
    \label{fig:emotion-decade}
\end{figure}

The inherent nature of the Billboard Hot 100, especially within recent years, acts as a testament to just how varied the emotions of songs can be within itself but across the charts holistically. Figure \ref{fig:emotion-decade} demonstrates this variety by displaying the distributions of the emotions in isolation across the dataset. We are able to see that the most prevalent emotion is 'joking' which comprises over 90\% of the dataset. We stress that the sentiment analysis model was trained using the SenWave data; it could be a possibility that the sentiment might be biased to predict 'joking' due to the inherent lexical structure of songs, which involves rhyming and the abstract construction of sentences. This hypothesis is further strengthened by the fact that there are no songs in the dataset identified as either 'pessimistic' or 'denial'. Potentially because the 'joking' sentiment creates more class imbalance, thereby preventing the prediction of the more extreme negative emotions. Despite this shortcoming, the sentiment analysis of these songs still provide some extremely valuable insights, especially in the context of all the songs which were identified to be either 'sad', 'annoyed' or 'anxious'. These emotions could still be considered to sit on the negative spectrum and reflect the increasing amount of abusive content that has crept into lyrics over the years. 

\begin{figure}[h!]
    \centering
    \includegraphics[width=\linewidth]{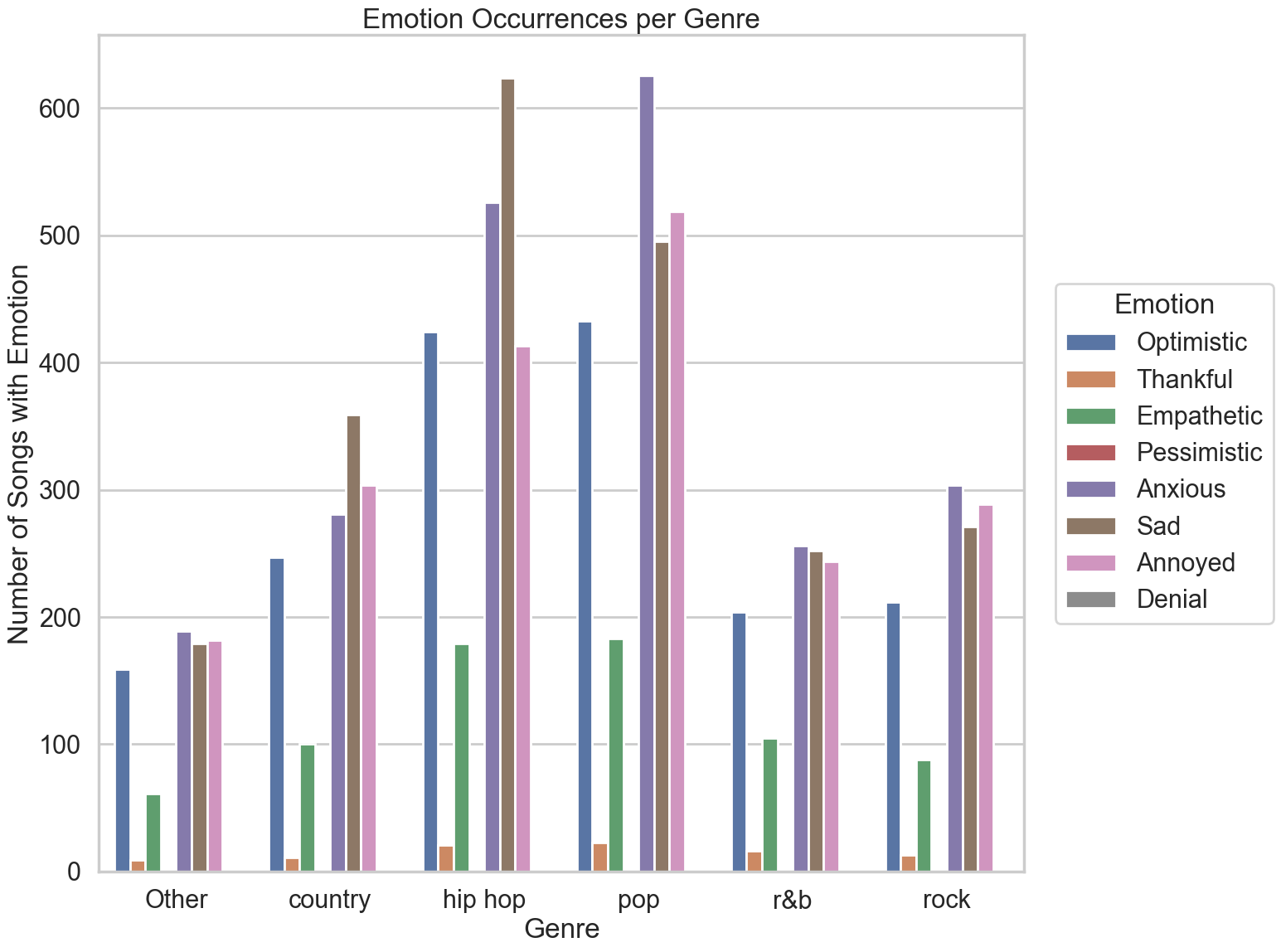}
    \caption{Emotion counts for different categories}
    \label{fig:emotion_counts_genre}
\end{figure}

\begin{figure}[h]
    \centering
    \includegraphics[width=1\linewidth]{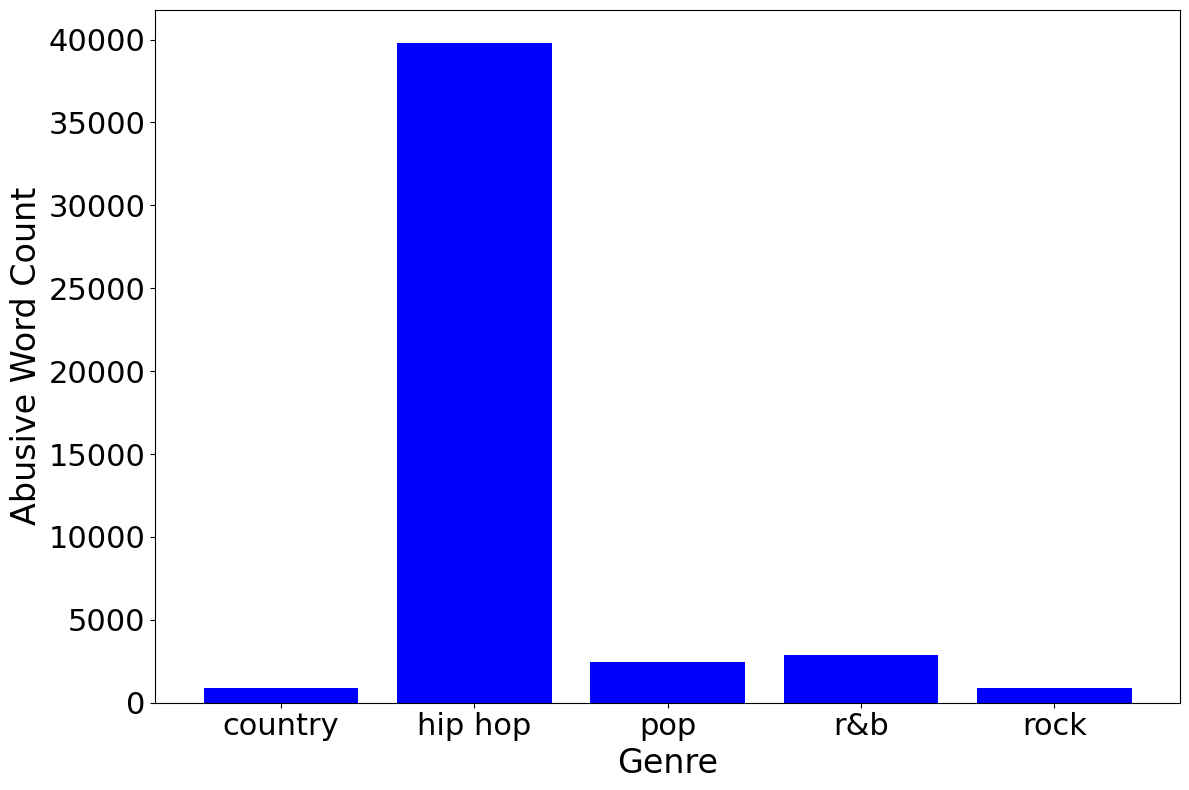}
    \caption{Bar chart of abusive word count within each genre}
    \label{fig:abusive_count_genre}
\end{figure}

A further partition of these sentiments into the primary genres of 'pop', 'hip hop', 'country', 'rock' and 'r\&b'  given in Figure \ref{fig:emotion_counts_genre}, also revealed interesting insights. We can observe that the distribution of the emotions within these genres seemed to be consistent overall, matching the overall distribution of the data. The primary differences between the genres occurred in the interchange of 'Anxious', 'Annoyed' and 'Sad' across the genres. However, the inspection of the relative proportion of songs that are classified as 'Optimistic' is inversely proportional to the amount of abusive words in the dataset by genre (Figure \ref{fig:abusive_count_genre}). We notice the Hip Hop genre having the fewest relative samples, followed by Rock, Country, R\&B and then Pop. The latter two genres had counts for 'Optimisic' that was equal to or greater than the 'Anxious' sentiment. The same conclusion is also supported by inspection of the converse 'Annoyed' sentiment, which tends to be directly proportional to the abusive word count. Once again, we find the Hip Hop genre having the most abusive word count, followed by R\&B, Pop, Rock and finally Country in that order.



\begin{figure}[h!]
    \centering
    \includegraphics[width=\linewidth]{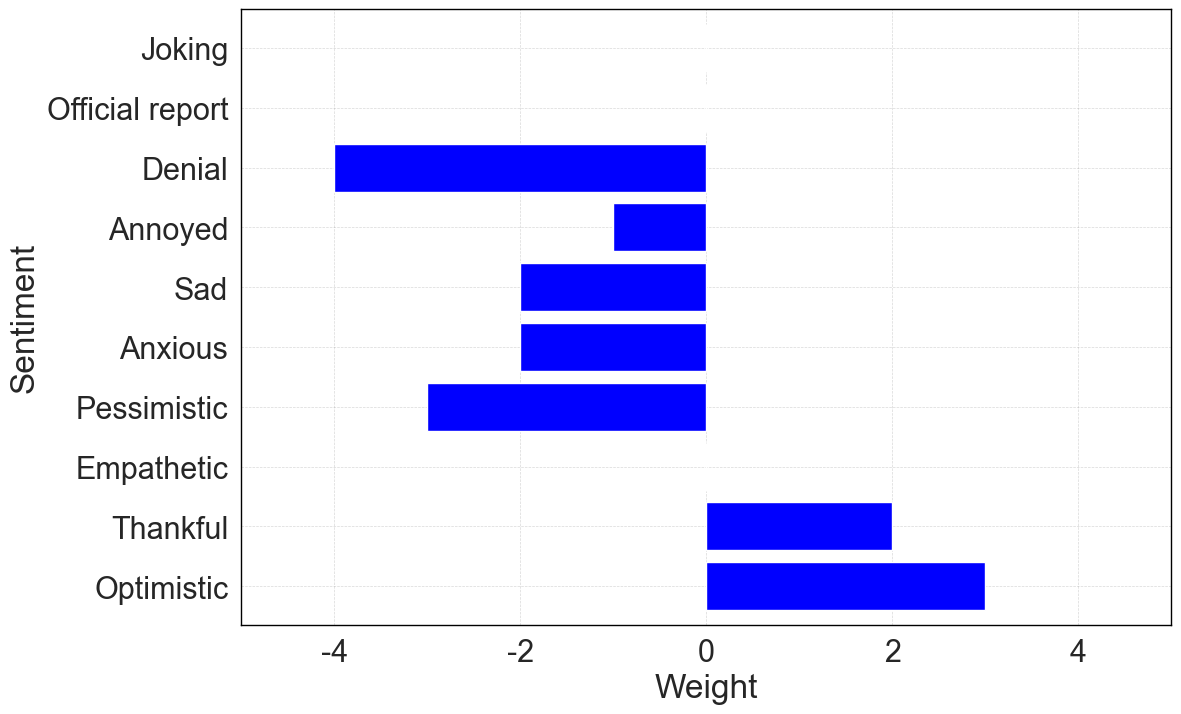}
    \caption{Bar chart of sentiment weights for custom polarity derivation}
    \label{fig:custom_sentiment_weights}
\end{figure}

\begin{figure}[h]
    \centering
    \includegraphics[width=1\linewidth]{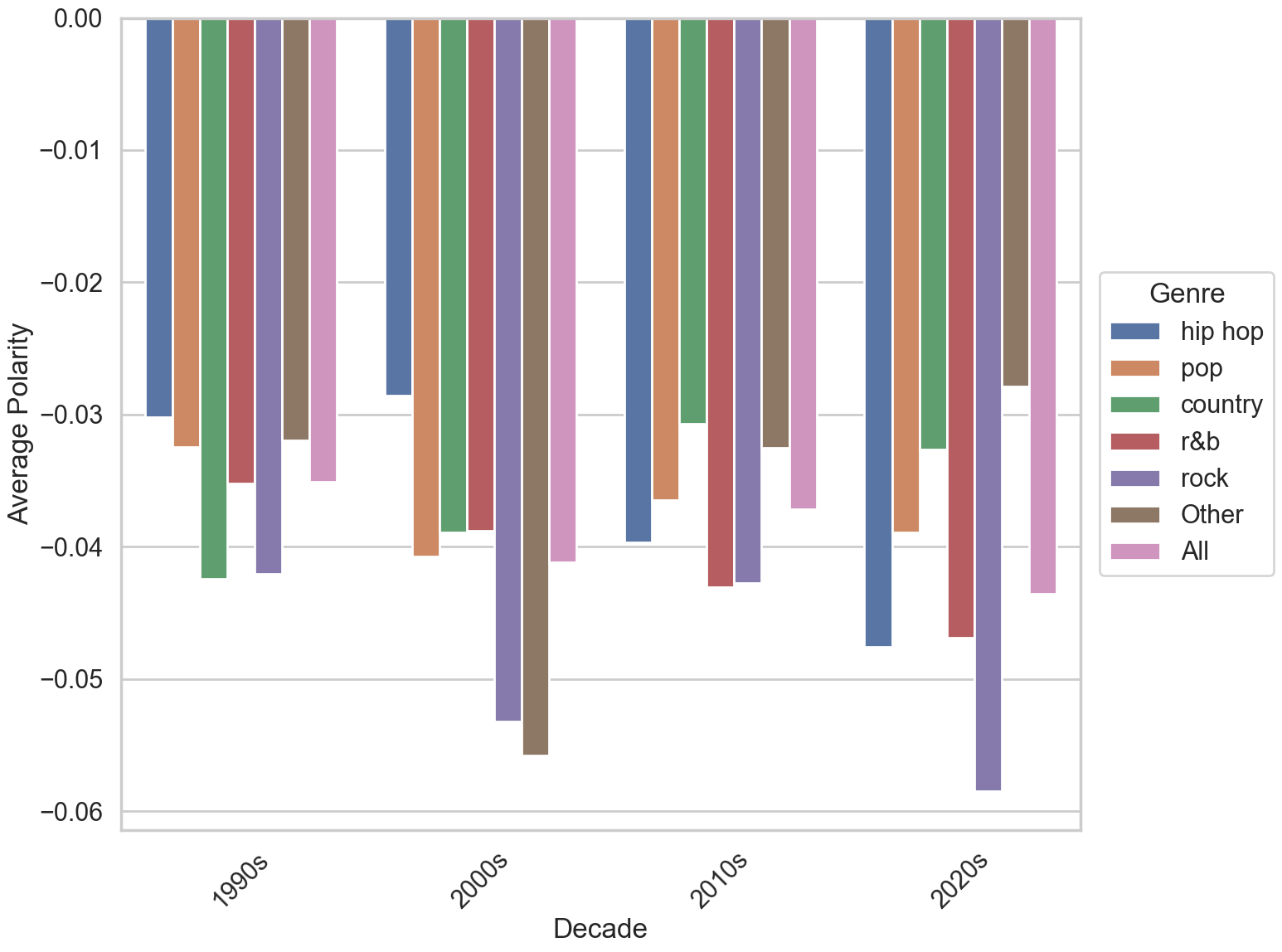}
    \caption{Average polarity score trend over years by genre}
    \label{fig:custom_polarity_genre_year}
\end{figure}

We adopt the strategy from Chandra et. al \cite{chandra2025longitudinal} for Hollywood movie dialogue sentiment analysis, where custom weights for the different emotions were developed. This enables us to derive a custom polarity score (Figure \ref{fig:custom_sentiment_weights}) by taking the weighted average to review sentiment expressed across the different years. 
Figure \ref{fig:custom_polarity_genre_year} presents sentiment polarity of songs partitioned into their charting date, showcasing that songs since the 1990s got progressively more negative before receding after 2005 until 2015, at which point it became worse. This implies that trends in popular songs regarding their use of abusive and negative content may be almost seasonal and cyclical.

A deeper delve into this graph by also partitioning it by the genre (Figure \ref{fig:custom_polarity_genre_year}) showcases that the individual genre's are exploring this cycle at different periods. The genres, such as Rock and Country, are quite similar to the overall trend, whilst Hip Hop and R\&B seem to consistently get more negative sentiments. Finally, Pop seems to be relatively stable in comparison to the other genres, with its amplitude and variance being the least out of the five genres.

\subsection{Abuse and Hate Speech Detection}

We used a look-up dictionary to count the number of abusive words to determine if a song was explicit or clean. 

\begin{figure}[htbp!]
    \centering
    \includegraphics[width=\linewidth]{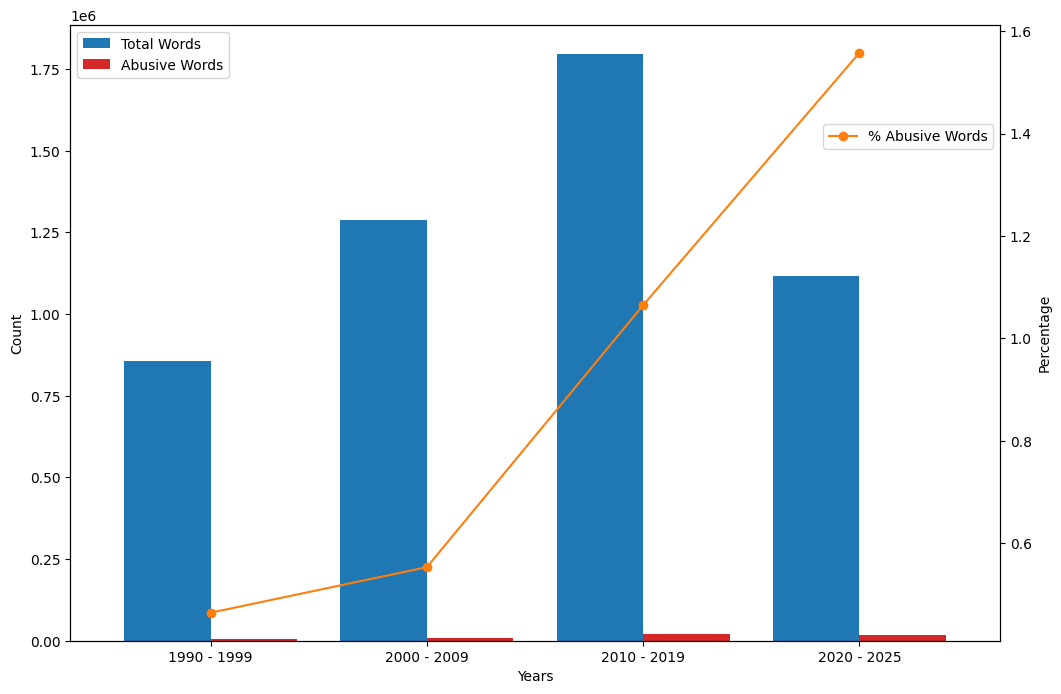}
    \caption{Total word count against abusive word count and relative percentage over years}
    \label{fig:total_words_vs_abusive_words_year_pct}
\end{figure}

Figure \ref{fig:total_words_vs_abusive_words_year_pct} explores the count of total words and abusive words in lyrical content over the year ranges. We observe that the total count seems to naturally increase with time, as expected with the growth of the music industry over the years. We can observe that the count of abusive words has also increased and an inspection of the relative percentage of abusive words concerning the total words over the years as seen in Figure \ref{fig:total_words_vs_abusive_words_year_pct} showcases that the proportion of these words are increasing.


\subsection{Case Studies}


To further validate the results of our analysis and determine the efficacy of using the polarity score to identify abusive content sequences within song lyrics, a few specific case studies were chosen in the five main genres of Pop, Hip Hop, Country, Rock and R\&B. We include the songs selected for each of these genres in their respective order.

\begin{figure*}[h!]
    \centering
    \includegraphics[width=\linewidth]{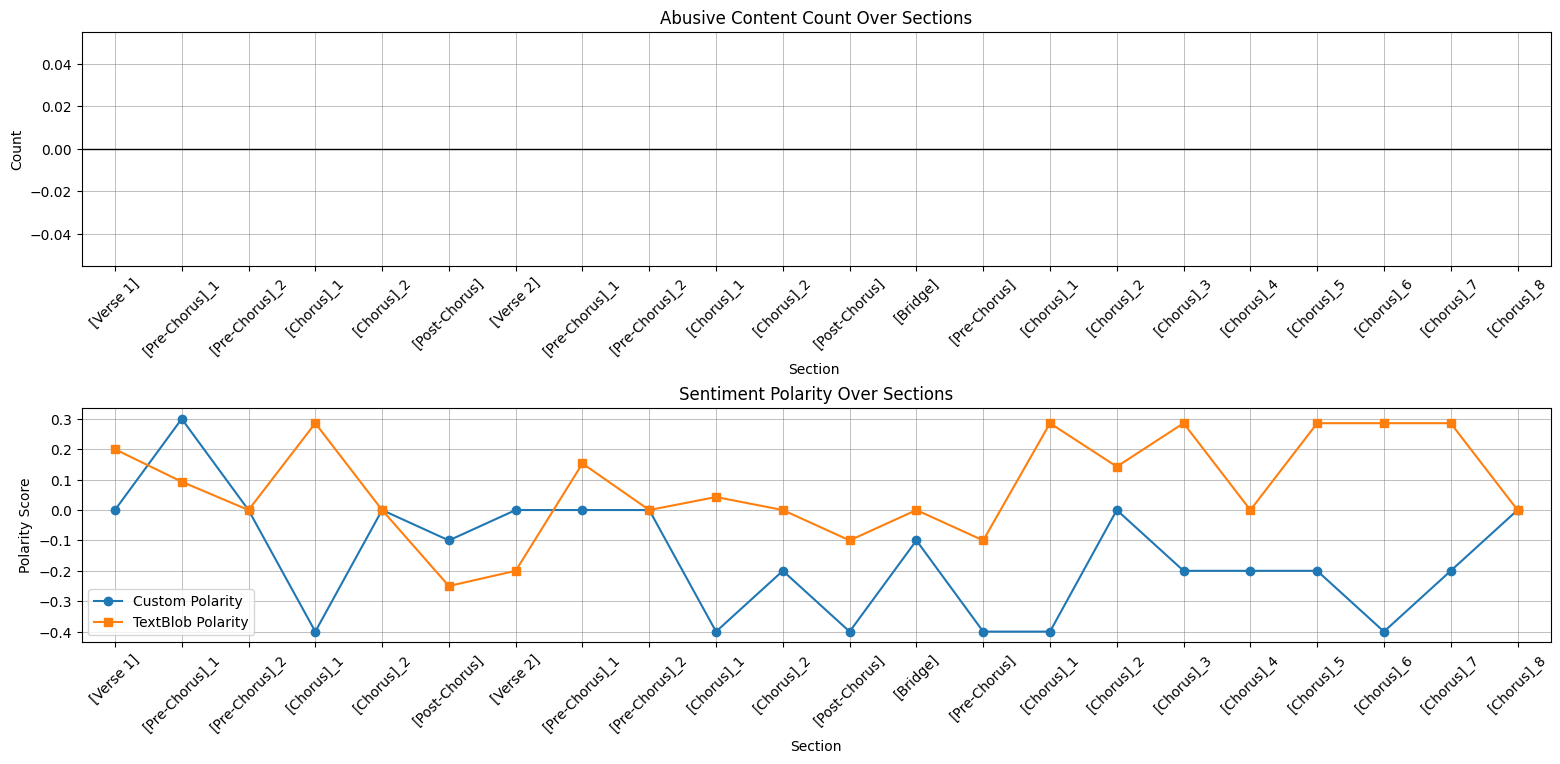}
    \caption{Longitudinal analysis of the sentiment \& abusive word count involved in Micheal Jackson's song 'Who Is It'}
    \label{fig:mj_who_is_it_analysis}
\end{figure*}

During sentiment analysis, we take into account the token limit for the model, and we partition the songs based on their sections (based on features), such as the chorus or a verse. If a verse exceeded the token limit, it would be further partitioned into smaller segments, as evidenced by the name of the section, which has an additional underscore after the section. Therefore, we analyse the song in its respective context, ensuring the validity of the sentiments across the span of the song. 
We present Michael Jackson's 'Who is it' (Figure \ref{fig:mj_who_is_it_analysis}) as a special case study due to the content of the song, which describes him spiralling down into a pit of depression after realising his partner had cheated on him with another man \cite{sahi2021michael}. The song paints an initial backdrop of true love and hope - evidenced in the lyrics of [Pre\_chorus]\_1: 

We highlight [Pre-Chorus] - 1:\textit{'and she promised me forever and a day we would live as one we made our vows we would live a life anew and she promised me in secret that she would love me for all time it is a promise'}

However, this image is quickly torn down once he finds out she was eloping with another man as showcased in the following in the first chunk of the chorus: 

\textit{'and it does not seem to matter and it does not seem right because the wheel has brought no fortune still I cry alone at night do not you judge of my composure because I am lying to myself'}

Figure \ref{fig:mj_who_is_it_analysis} presents the inspection of the polarity score for both the custom weights metric and TextBlob polarity for the song. We find that the initial hopeful backdrop presented in Pre-Chorus 1 is classified to be 'Optimistic' with a custom polarity score of 0.3. On the other hand, the lyric as described earlier in Chorus 1 had sentiments of 'anxious' and 'sad' with the custom polarity nosediving straight to -0.4 in response to the dark and depressive content. Surprisingly, even though the content of the song could be considered abusive due to the dark and mature themes present within the lyrics, it would not be suitable for children. The actual count of abusive words used in the song equalled zero, as revealed in Figure \ref{fig:mj_who_is_it_analysis}, showcasing how easy it is to discuss abusive content without explicit reference to expletives.

\begin{figure*}[h!]
    \centering
    \includegraphics[width=\linewidth]{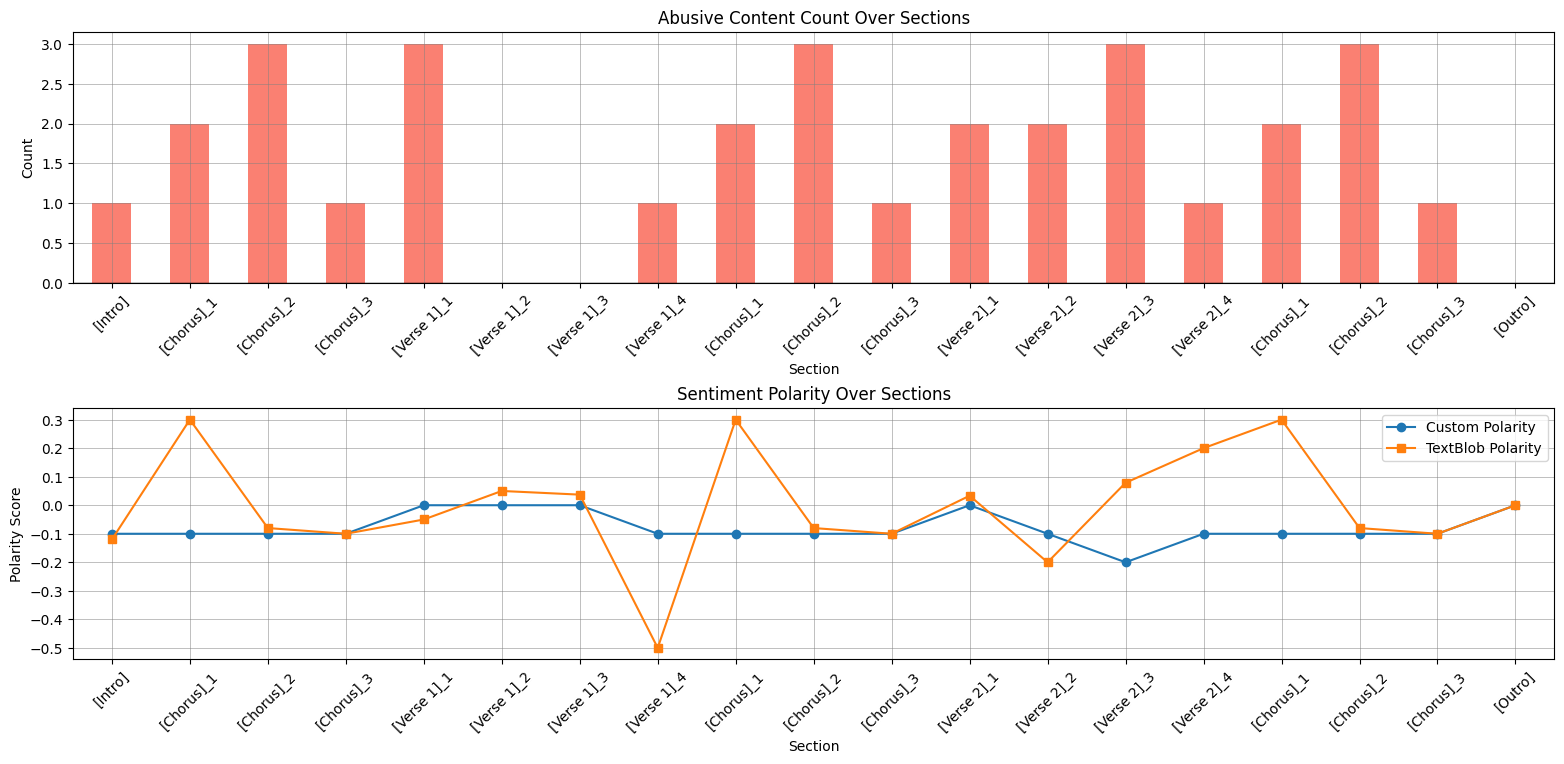}
    \caption{Longitudinal analysis of the sentiment \& abusive word count involved in Pooh Shiesty's song 'Back in Blood'}
    \label{fig:ps_back_in_blood_analysis}
\end{figure*}

In the case of Hip Hop, the analysis of lyrics and songs is nuanced, as there are situations where expletives are used, but also situations where colloquial terms are used that would significantly alter the sentiment of the sentence. This is effectively demonstrated in some of the lyrics and associated polarity scores for the lyrics of Pooh Shiesty's song featuring Lil Durk "Back In Blood" (Figure \ref{fig:ps_back_in_blood_analysis}).
We highlight [Verse 1] - 3:\textit{"Glock keep my door unlocked and stop I like gettin on feet park the car blrrrd we gettin up close do him dirty i not showin love eleven thousand all ones left my right pocket in the club these blue"}. This extract is the third chunk that was processed from the first verse. For context, the ad-lib "blrrd" is a signature phrase that Pooh Shiesty uses across and within his songs quite extensively, even though it doesn't hold any significant meaning. In addition to this, the euphemism of "do him dirty" to describe him grievously injuring someone also poses an issue for the model to accurately determine the custom polarity of this verse, as evidenced by its score of zero (Figure \ref{fig:ps_back_in_blood_analysis}). The TextBlob polarity has an even worse score in comparison with its prediction of 0.03, which indicates that the verse has an overall positive sentiment.

Despite the existence of these euphemisms and slang words, the weighted sentiment score is still able to effectively capture abusive content. Inspecting [Verse 2] - 3, the associated sentiment score is -0.2; the corresponding lyrics for this section are the following. We highlight [Verse 2] - 3: \textit{roy pop up out that cut with that new glock i wish he would run his ass playin bi*ch I am really icy pooh shiesty that is my dawg but pooh you know I am really shiesty you told all them o t n****s}. Within these lyrics, there is a clear reference to slang as well - the word 'glock' referencing a specific type of gun, which would cause the victim to "run" for his life. Furthermore, the word 'shiesty' is used as an adjective to describe someone as "shady or untrustworthy" thereby insinuating and suggesting clear violence. Despite the involvement of slang within the lyrics, the sentiment is correctly assessed and determined.

\begin{figure*}[h!]
    \centering
    \includegraphics[width=\linewidth]{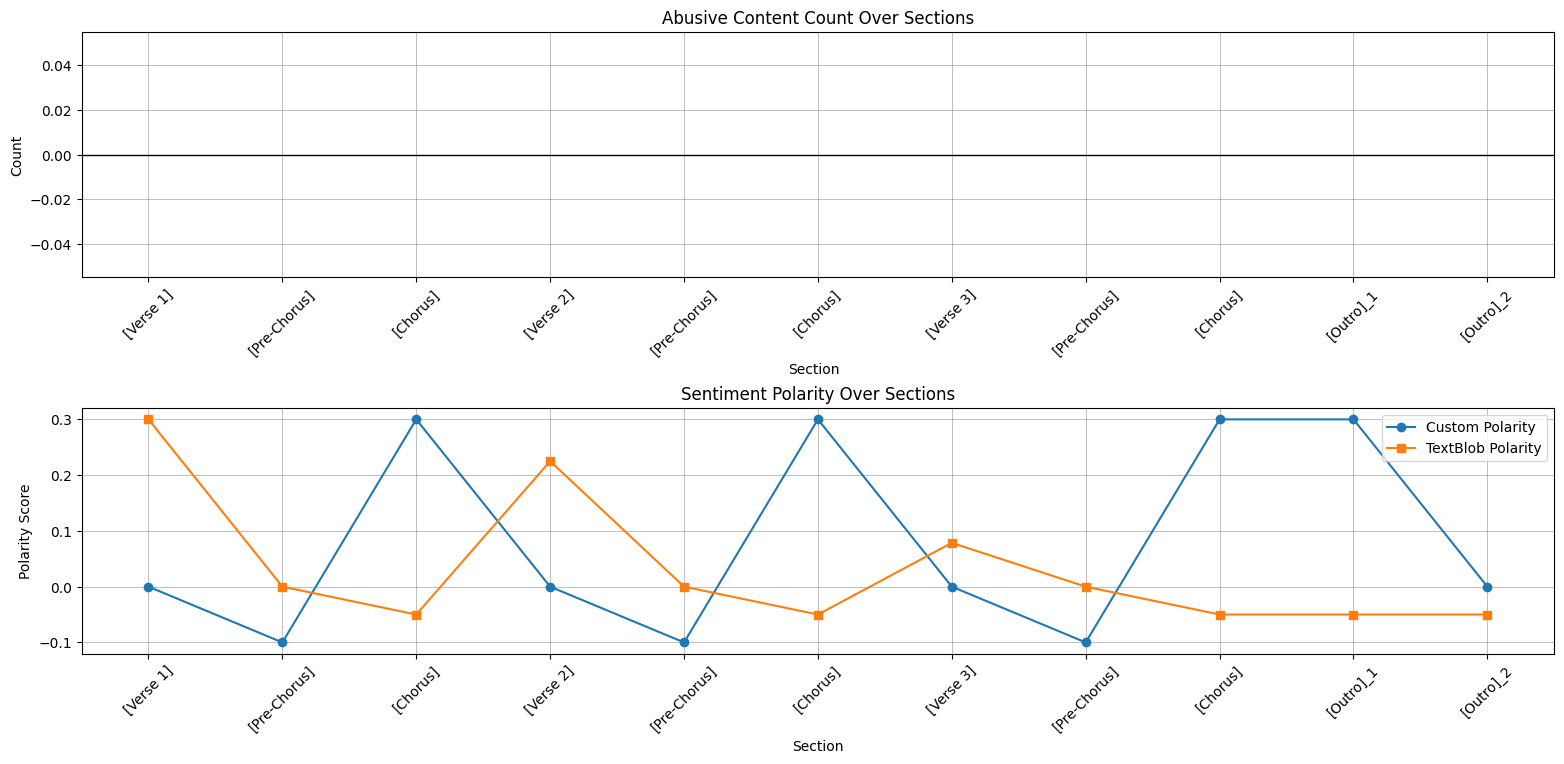}
    \caption{Longitudinal analysis of the sentiment \& abusive word count involved in Mary Chapin Carpenter's song 'Passionate Kisses'}
    \label{fig:mcc_passionate_kisses_analysis}
\end{figure*}

Unlike Micheal Jackson's 'Who is it', which references emotionally charged content that could be considered inappropriate for younger audiences, songs from hip hop tend to be much more overt with their direct use of expletives in their lyrics. At the same time, however, the use of euphemisms and slang is much more apparent in this genre, which then requires careful inspection of the lyrical content to discern the real meaning and sentiment of the songs. Similar to the Pop genre, Country songs tend to be rather mellow, with relatively few songs that contain overtly explicit and abusive content. This is reflected by the percentage of explicit songs in the dataset altogether, which is approximately 1.6\%, i.e (25/1519). "Passionate Kisses" by Mary Chapin Carpenter (Figure \ref{fig:mcc_passionate_kisses_analysis}) is a song that is considered non-explicit as per Spotify's guidelines, which is effectively reflected in the analysis of its lyrical content as well. There is no use of words that are deemed to be explicit (Figure \ref{fig:mcc_passionate_kisses_abusive_count}), the song describes the pining of the artist to experience "passionate kisses" and a blissful life overall. 





\begin{figure*}[h!]
    \centering
    \includegraphics[width=\linewidth]{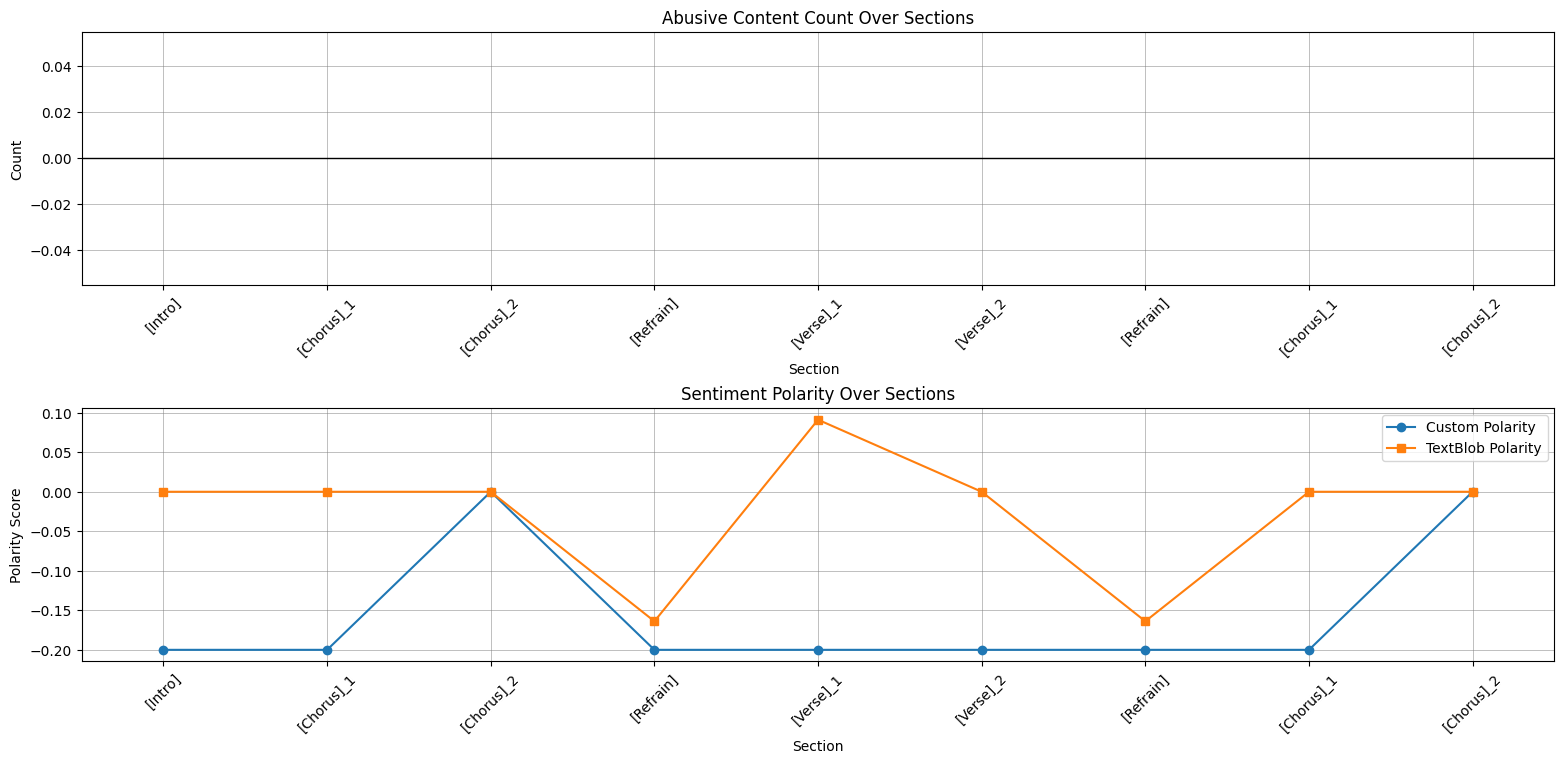}
    \caption{Longitudinal analysis of the sentiment \& abusive word count involved in Future's song 'XanaX Damage'}
    \label{fig:future_xanax_damage_analysis}
\end{figure*}

Although the R\&B genre is known for its soulful tunes and melodies and generally clean and non-abusive lyrics, there are instances of songs referencing abusive content. "XanaX Damage" by Future (Figure \ref{fig:future_xanax_damage_analysis}) is a song that discusses the impact of Xanax, a prescription drug, on the artist and his dependency on the substance. Even though the song lyrics do not contain any expletives (Figure \ref{fig:future_xanax_damage_analysis}), it is evident that the song covers explicit material - reflected by Spotify's labelling and further cemented by the sentiment analysis of its lyrics within the Chorus and first chunk of the only verse in the song which both had a custom sentiment score of 0.2 (Figure \ref{fig:future_xanax_damage_analysis}).




Finally, the rock genre in of itself has an extremely wide and varied range of emotional expression when it comes to the lyrical content of the songs within it. Sub-genres such as Heavy Metal have been described as "overwhelmingly dominated" by "ugly and unhappy" emotions that tend to express "no hope" for the future by Jeffrey Arnett - a Psychology professor at the Clark University in Massachusetts. On the other hand, the intersection of Pop and Rock in the adequately termed "Pop Rock" sub-genre tend to be much more lighter and happier in lyrical content.

\begin{figure*}[h!]
    \centering
    \includegraphics[width=\linewidth]{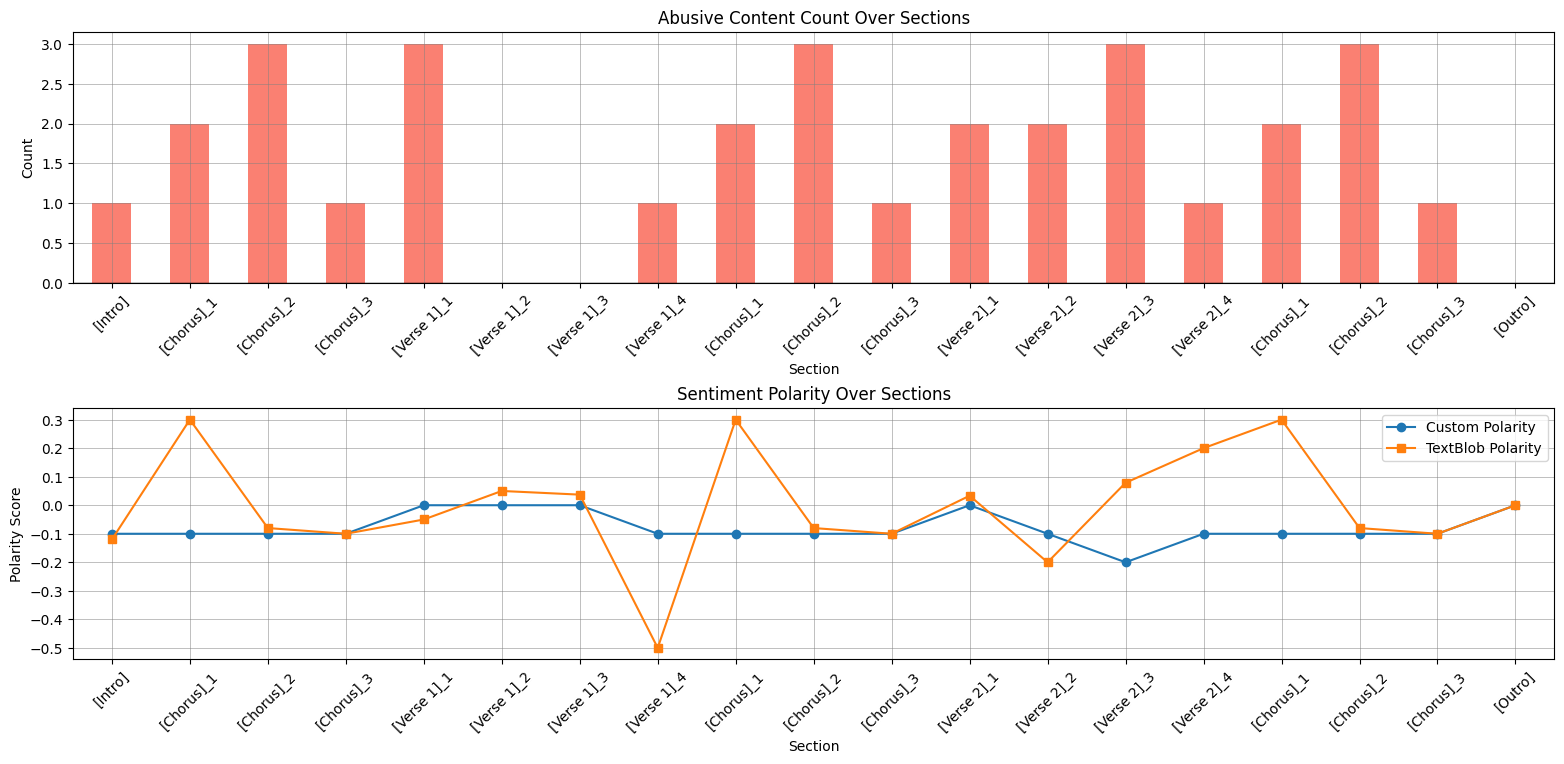}
    \caption{Longitudinal analysis of the sentiment and abusive word count involved in Linkin Park's song 'In The End'.}
    \label{fig:lp_in_the_end_analysis}
\end{figure*}

Linkin Park's "In The End" (Figure \ref{fig:lp_in_the_end_analysis}) embraces the nihilistic motif described earlier, describing how all effort is meaningless "in the end". Once again, despite the fact that there are no expletives used within the lyrical content of the song, the material it discusses is naturally not appropriate for young children. This is reflected by the pre-chorus and second chunk of the first verse, which both had polarity scores of -0.2 and emotions of 'Sad' and 'Anxious' respectively (Figure \ref{fig:lp_in_the_end_analysis}). Note the [Pre-chorus]:\textit{"I kept everything inside and even though I tried it all fell apart, what it meant to me will eventually be a memory of a time when I tried so hard"}. We highlight  [Verse 1] -2:\textit{"swings watch it count down to the end of the day the clock ticks life away it is so unreal did not look out below watch the time go right out the window tryin to hold on did not even know I wasted."} Similar to Michael Jackson's "Who Is It" and Future's "XanaX Damage", even though the song has no explicit reference to expletives and is also considered clean as per Spotify's standards (Figure \ref{fig:lp_in_the_end_analysis}). The content of the lyrics reflects and discusses darker themes that are not appropriate for children.

\section{Discussion}

In this study, we assessed the efficacy of various models and embeddings for predicting the explicit and non-explicit nature of songs. 

One limitation of our study is that the classification of songs as explicit or non-explicit relies on labels obtained from the Spotify API. Since Spotify's classification of songs as explicit may not always align perfectly with the linguistic criteria our model uses, there is a risk of bias in our training data. This could lead to our model disproportionately predicting songs as explicit simply because they are labelled as such in the Spotify database, regardless of the actual linguistic content. Future studies could mitigate this limitation by incorporating multiple sources of classification labels or implementing human validation of song content to ensure more accurate training data.

Our results have significant implications for the music industry, particularly in the areas of content moderation and recommendation systems. The models developed in this study can enhance the accuracy and efficiency of filtering explicit content on streaming platforms, ensuring that users receive appropriate song recommendations. Additionally, these models can be integrated into parental control systems to better safeguard young listeners from explicit lyrics, providing a more tailored and secure music listening experience. By attempting to address these knowledge gaps in the current research space, we aim to provide value by equipping policymakers, parents and other relevant parties with the knowledge and tools required to effectively identify and regulate content that can be considered inappropriate for vulnerable groups such as children and adolescents. 

Future work could extend this research by incorporating additional contextual factors, such as music genre and regional differences, to refine the classification models further. Moreover, integrating more advanced LLMs and exploring the impact of explicit content on listener behaviour and mental health could provide deeper insights into the broader implications of lyrical content in popular music. Overall, this study highlights the critical role of technology in understanding and addressing the evolving landscape of media content in the digital age.

\section{Conclusion}

In this study, we have explored the use of deep learning methods and NLP to perform a longitudinal analysis of abusive and inappropriate content in Billboard Music Charts over the past seven decades. We utilised various advanced NLP techniques, including n-gram analysis, sentiment analysis, and abusive content detection, to identify and categorise explicit lyrics. Our analysis has demonstrated the significant rise in explicit content within popular music, particularly from 1990 onwards. This trend highlights the increasing prevalence of songs with lyrics containing profane, sexually explicit, and otherwise inappropriate language.
 Our results showcased the ability of these models to capture nuanced patterns in lyrical content, reflecting shifts in societal norms and language use over time.

The results underscore the importance of developing sophisticated automated content filtering systems that can adapt to evolving language and cultural contexts. By providing a robust framework for identifying explicit content in music, this study contributes to the ongoing efforts to safeguard young audiences from potentially harmful material. The insights gained from this research can aid parents, educators, and policymakers in making informed decisions regarding children's musical consumption, fostering a safer and more appropriate media environment.

\bibliographystyle{elsarticle-num} 
\bibliography{cas-refs, references}

\section*{Appendix}

\subsection{Granular View on Song Chunk Analysis}

\begin{table*}
    \centering
    \begin{adjustbox}{width=\textwidth}
    \small
    \begin{tabular}{|p{9cm}|p{2cm}|p{1.5cm}|p{1.2cm}|p{1.2cm}|p{1.5cm}|}
        \hline
        \textbf{Chunk} & \textbf{Section} & \textbf{Emotions} & \textbf{Abusive Words} & \textbf{Polarity Score} & \textbf{TextBlob Polarity} \\
        \hline
        i gave her money i gave her time i gave her everything inside one heart could find i gave her passion my very soul i gave her promises and secrets so untold & Verse 1  & N/A & 0 & 0 & 0.2 \\
        \hline
        and she promised me forever and a day we would live as one we made our vows we would live a life anew and she promised me in secret that she would love me for all time it is a promise so untrue tell & Pre\_chorus 1  & Optimistic & 0 & 0.3 & 0.09\\
        \hline
         me what will i do & Pre-Chorus\_2 & N/A & 0 &  0 & 0 \\
         \hline
         and it does not seem to matter and it does not seem right because the wheel has brought no fortune still i cry alone at night do not you judge of my composure because i am lying to myself and the & [Chorus]\_1 & Anxious, Sad & 0 & -0.4 & 0.28 \\
         \hline
         reason why she left me did she find in someone else & [Chorus]\_2 & N/A & 0 & 0 & 0\\
         \hline
         who is it is it a friend of mine who is it is it my brother who is it somebody hurt my soul now who is it i cannot take this stuff no more & [Post-Chorus] & Annoyed & 0 & -0.1 & -0.25 \\
         \hline
         i am the damned i am the dead i am the agony inside dying head this is injustice woe unto thee i pray this punishment would have mercy on me & [Versfirste 2] & Empathetic & 0 & 0 & -0.2 \\
         \hline
         and she promised me forever that we would live our life as one we made our vows we would live a love so true it seems that she has left me for such reasons unexplained i need to find the truth but & Pre-Chorus\_1 & Joking & 0 &  0 & 0.15 \\
         \hline
         see what will i do & Pre-Chorus\_2  & N/A & 0 & 0 & 0\\
         \hline
         and it does not seem to matter and it does not seem right because the wheel has brought no fortune still i cry alone at night do not you judge of my composure because i am bothered everyday and she & [Chorus]\_1 & Anxious, Sad & 0 & -0.4 & 0.04\\
         \hline
         did not leave a letter she just up and ran away & [Chorus]\_2 & Sad & 0 & -0.2 & 0\\
         \hline
         who is it it is a friend of mine who is it is it my brother who is it somebody hurt my soul now who is it i cannot take it because i am lonely & [Chorus]\_3 & Anxious, Sad & 0 & -0.4 & -0.1\\
         \hline
         do not try to pick me do not try quit it i never was we never were just will not stop do not bother me do not bother me & [Chorus]\_4 & Annoyed & 0 & -0.1 & -0.1\\
         \hline
         who is it it is a friend of mine who is it to me i am bothered who is it somebody hurt my soul now who is it i cannot take it because i am lonely & [Chorus]\_5 & Anxious, Sad & 0 & -0.4 & -0.1\\
         \hline
         and it does not seem to matter and it does not seem right because the wheel has brought no fortune still i cry alone at night do not you judge of my composure because i am lying to myself and the & [Chorus]\_6 & Anxious, Sad & 0 & -0.4 & 0.28\\
         \hline
         reason why she left me did she find in someone else and it does not seem to matter and it does not seem right because the wheel has brought no fortune still i cry alone at night do not you judge of & [Chorus]\_7 & N/A & 0 & 0 & 0.14\\
         \hline
         my composure because i am bothered every day and she did not leave a letter she just upped and ran away and it does not seem to matter and it does not seem right because the wheel has brought no & [Chorus]\_8 & Anxious & 0 & -0.2 & 0\\
         \hline
        fortune still i cry alone at night do not you judge of my composure because i am lying to myself and the reason why she left me did she find in someone else and it does not seem to matter and it does & [Chorus]\_9 & Anxious & 0 & -0.2 & 0\\
         \hline
         not seem right because the wheel has brought no fortune still i cry alone at night do not you judge of my composure because i am bothered every day and she did not leave a letter she just upped and & [Chorus]\_10 & Sad & 0 & -0.2 & 0.28\\
         \hline
         ran away and it does not seem to matter and it does not seem right because the wheel has brought no fortune still i cry alone at night do not you judge of my composure because i am bothered every day & [Chorus]\_11 & Anxious, Sad & 0 & -0.4 & 0.28\\
         \hline
         and she did not leave a letter she just upped and ran away and it does not seem to matter and it does not seem right because the wheel has brought no fortune still i cry alone at night do not you & [Chorus]\_12 & Anxious & 0 & -0.2 & 0.28\\
         \hline
         judge of my composure because i am lying to myself and the reason why she left me did she find in someone else 25 & [Chorus]\_13 & N/A & 0 & 0 & 0\\
         \hline
    \end{tabular}
    \end{adjustbox}
    \caption{Comprehensive view on sentiment and abuse detection for Michael Jackon's song "Who is it"}
    \label{tab:mj_who_is_it_chunks}
\end{table*}

\begin{table*}
    \centering
    \begin{adjustbox}{width=\textwidth}
    \small
    \begin{tabular}{|p{9cm}|p{2cm}|p{1.5cm}|p{1.2cm}|p{1.2cm}|p{1.5cm}|}
        \hline
        \textbf{Chunk} & \textbf{Section} & \textbf{Emotions} & \textbf{Abusive Words} & \textbf{Polarity Score} & \textbf{TextBlob Polarity} \\
        \hline
        smurk big blrrrd huh i do not know why he want something back from me shit shit you gotta get this shit in blood homie & Intro  & Annoyed & 1 & -0.1 & -0.12 \\
        \hline
        bitch i got my own fire do not need security in the club all that woofin on the net nigga i thought you was a thug they not got nowhere to go i shot up everywhere they was yeah you know who took that & Chorus\_1  & Annoyed & 2 & -0.1 & 0.3 \\
        \hline
        shit from you come get it back in blood bitch come get it back in blood we not mask up no dodgin wrecks niggas know who it was extortin shit just like the 80s want something back get it in blood yeah & Chorus\_2  & Annoyed & 3 & -0.1 & -0.08 \\
        \hline
        you know who took that shit from you come get it back in blood & Chorus\_3  & Annoyed & 1 & -0.1 & -0.1 \\
        \hline
        if your nigga killer not dead you should not wear no r i p shirt we had three hundred shots up in the car before you picked up durk you niggas who not got shit goin go grab a glizzy get alert shiesty & Verse 1\_1  & Joking & 3 & 0 & -0.05 \\
        \hline
        g post r i p and reason he in the dirt you gotta know i go too far it is two o s up on this hundred one of em might stand for o block bout twenty some shots left up in the k fifteen still in the & Verse 1\_2  & Joking & 0 & 0 & 0.05 \\
        \hline
        glock keep my door unlocked and stop i like gettin on feet park the car blrrrd we gettin up close do him dirty i not showin love eleven thousand all ones left my right pocket in the club these blue & Verse 1\_3  & Joking & 0 & 0 & 0.03 \\
        \hline
        faces up on me dirty i went and got it out the mud if i took something get it in blood i do not give a fuck what we was & Verse 1\_4  & Annoyed & 1 & -0.1 & -0.5 \\
        \hline
        bitch i got my own fire do not need security in the club all that woofin on the net nigga i thought you was a thug they not got nowhere to go i shot up everywhere they was yeah you know who took that & Chorus\_1  & Annoyed & 2 & -0.1 & 0.3 \\
        \hline
       shit from you come get it back in blood bitch come get it back in blood we not mask up no dodgin wrecks niggas know who it was extortin shit just like the 80s want something back get it in blood yeah & Chorus\_2  & Annoyed & 3 & -0.1 & -0.08 \\
        \hline
        you know who took that shit from you come get it back in blood & Chorus\_3  & Annoyed & 1 & -0.1 & -0.1 \\
        \hline
        killed your mans you keep on talkin better get that shit in blood give my shawty and them a dub then they gon walk inside this club hit his lil ass with that switch i bet that switch switch up his & [Verse 2]\_1  & Joking & 2 & 0 & 0.03 \\
        \hline
        nerves fuck the opps inside my city lil bro put them in the mud you cannot come back to your hood huh he was dissin on my cousin now his ass all in that wood huh book his ass i wish he would come v & [Verse 2]\_2  & Annoyed & 2 & -0.1 & -0.2 \\
        \hline
        roy pop up out that cut with that new glock i wish he would run his ass playin bitch i am really icy pooh shiesty that is my dawg but pooh you know i am really shiesty you told all them o t niggas & [Verse 2]\_3  & Sad & 3 & -0.2 & 0.07 \\
        \hline
        that you really slide tell the truth about your gang bitch they really dyin & [Verse 2]\_4 & Annoyed & 1 & -0.1 & 0.2 \\
        \hline
        bitch i got my own fire do not need security in the club all that woofin on the net nigga i thought you was a thug they not got nowhere to go i shot up everywhere they was yeah you know who took that & [Chorus]\_1 & Annoyed & 2 & -0.1 & 0.3 \\
        \hline
        shit from you come get it back in blood bitch come get it back in blood we not mask up no dodgin wrecks niggas know who it was extortin shit just like the 80s want something back get it in blood yeah & [Chorus]\_2 & Annoyed & 3 & -0.1 & -0.08 \\
        \hline
        you know who took that shit from you come get it back in blood blrrrd & [Chorus]\_3 & Annoyed & 1 & -0.1 & -0.1 \\
        \hline
        come get it back in blood 58 & [Outro] & N/A & 0 & 0 & 0\\
        \hline
        
    \end{tabular}
    \end{adjustbox}
    \caption{Comprehensive view on sentiment and abuse detection for Pooh Sheisty's song "Back in Blood"}
    \label{tab:ps_back_in_blood_chunks}
\end{table*}

\begin{table*}
    \centering
    \begin{adjustbox}{width=\textwidth}
    \small
    \begin{tabular}{|p{9cm}|p{2cm}|p{1.5cm}|p{1.2cm}|p{1.2cm}|p{1.5cm}|}
        \hline
        \textbf{Chunk} & \textbf{Section} & \textbf{Emotions} & \textbf{Abusive Words} & \textbf{Polarity Score} & \textbf{TextBlob Polarity} \\
        \hline
        is it too much to ask i want a comfortable bed that will not hurt my back food to fill me up and warm clothes and all that stuff & [Verse 1] & Annoyed & 0 & 0 & 0.3 \\
        \hline
        should not i have this should not i have this should not i have all of this and & [Pre-Chorus] & Annoyed & 0 & -0.1 & 0 \\
        \hline
        passionate kisses passionate kisses passionate kisses from you & [Chorus] & Optimistic, Joking & 0 & 0.3 & -0.05 \\
        \hline
        is it too much to demand i want a full house and a rock and roll band pens that will not run out of ink and cool quiet and time to think & [Verse 2] & Joking & 0 & 0 & 0.22 \\
        \hline
        should not i have this should not i have this should not i have all of this and & [Pre-Chorus] & Annoyed & 0 & -0.1 & 0 \\
        \hline
        passionate kisses passionate kisses passionate kisses from you & [Chorus] & Optimistic, Joking & 0 & 0.3 & -0.05 \\
        \hline
        do i want too much am i going overboard to want that touch i shout it out to the night give me what i deserve cause it is my right & [Verse 3] & Joking & 0 & 0 & 0.07 \\
        \hline
        should not i have this should not i have this should not i have all of this and & [Pre-Chorus] & Annoyed & 0 & -0.1 & 0 \\
        \hline
        passionate kisses passionate kisses passionate kisses from you & [Chorus] & Optimistic, Joking & 0 & 0.3 & -0.05 \\
        \hline
        passionate kisses passionate kisses from you passionate kisses passionate kisses from you passionate kisses passionate kisses from you passionate kisses passionate kisses from you passionate kisses & [Outro]\_1 & Optimistic, Joking & 0 & 0.3 & -0.05 \\
        \hline
        passionate kisses from you & [Outro]\_2 & Joking & 0 & 0 & -0.05 \\
        \hline
    \end{tabular}
    \end{adjustbox}
    \caption{Comprehensive view on sentiment and abuse detection for  Mary Chapin Carpenter's song "Passionate Kisses"}
    \label{tab:ps_back_in_blood_chunks}
\end{table*}

\begin{table*}
    \centering
    \begin{adjustbox}{width=\textwidth}
    \small
    \begin{tabular}{|p{9cm}|p{2cm}|p{1.5cm}|p{1.2cm}|p{1.2cm}|p{1.5cm}|}
        \hline
        \textbf{Chunk} & \textbf{Section} & \textbf{Emotions} & \textbf{Abusive Words} & \textbf{Polarity Score} & \textbf{TextBlob Polarity} \\
        \hline
        blood blood blood xanax dreams xanax covered in blood & [Intro] & Anxious & 0 & -0.2 & 0 \\
        \hline
        baby when the sun is out it is like i am not myself xanax doin damage make it night so i can handle you baby when the moon is out i finally know myself xanny pills then i feel my body startin to give & [Chorus]\_1 & Sad & 0 & -0.2 & 0 \\
        \hline
        up yeah & [Chorus]\_2 & Joking & 0 & 0 & 0 \\
        \hline
        i only call you when i am faded your arms around me come and save me i only want you to have my baby when i am drunk and i am down and depressed i just need to confess to you & [Refrain] & Sad & 0 & -0.2 & -0.16 \\
        \hline
        baby if i want you then i know there is somethin wrong i do not mean to ruin all the times we had alone but i am not my best with you i am so depressed with you but it is so hard i do not think i can & [Verse]\_1 & Sad & 0 & -0.2 & 0.09 \\
        \hline
        exist without you & [Verse]\_2 & Sad & 0 & -0.2 & 0 \\
        \hline
        i only call you when i am faded your arms around me come and save me i only want you to have my baby when i am drunk and i am down and depressed i just need to confess to you & [Refrain] & Sad & 0 & -0.2 & -0.16 \\
        \hline
        baby when the sun is out it is like i am not myself xanax doin damage make it night so i can handle you baby when the moon is out i finally know myself xanny pills then i feel my body startin to give & [Chorus]\_1 & Sad & 0 & -0.2 & 0 \\
        \hline
        up yeah & [Chorus]\_2 & Joking & 0 & 0 & 0 \\
        \hline
    \end{tabular}
    \end{adjustbox}
    \caption{Comprehensive view on sentiment and abuse detection for  Future's song "Xanax Damage"}
    \label{tab:ft_xanax_dmg_chunks}
\end{table*}

\begin{table*}
    \centering
    \begin{adjustbox}{width=\textwidth}
    \small
    \begin{tabular}{|p{9cm}|p{2cm}|p{1.5cm}|p{1.2cm}|p{1.2cm}|p{1.5cm}|}
        \hline
        \textbf{Chunk} & \textbf{Section} & \textbf{Emotions} & \textbf{Abusive Words} & \textbf{Polarity Score} & \textbf{TextBlob Polarity} \\
        \hline
        it starts with one & [Intro] & N/A & 0 & 0 & 0 \\
        \hline
        one thing i do not know why it does not even matter how hard you try keep that in mind i designed this rhyme to explain in due time all i know time is a valuable thing watch it fly by as the pendulum & [Verse 1]\_1 & Joking & 0 & 0 & 0.12 \\
        \hline
        swings watch it count down to the end of the day the clock ticks life away it is so unreal did not look out below watch the time go right out the window tryin to hold on d did not even know i wasted & [Verse 1]\_2 & Anxious & 0 & -0.2 & -0.02 \\
        \hline
        it all just to watch you go & [Verse 1]\_3 & Joking & 0 & 0 & 0 \\
        \hline
        i kept everything inside and even though i tried it all fell apart what it meant to me will eventually be a memory of a time when i tried so hard & [Pre-Chorus] & Sad & 0 & -0.2 & -0.29 \\
        \hline
        i tried so hard and got so far but in the end it does not even matter i had to fall to lose it all but in the end it does not even matter & [Chorus] & N/A & 0 & -0.2 & -0.29 \\
        \hline
       one thing i do not know why it does not even matter how hard you try keep that in mind i designed this rhyme to remind myself how i tried so hard in spite of the way you were mockin me actin like i & [Verse 2]\_1 & Annoyed, Joking & 0 & -0.1 & -0.29 \\
        \hline
        was part of your property remembering all the times you fought with me i am surprised it got so far things are not the way they were before you would not even recognize me anymore not that you knew & [Verse 2]\_2 & Anxious & 0 & -0.2 & 0.1 \\
        \hline
        me back then but it all comes back to me in the end & [Verse 2]\_3 & Sad & 0 & -0.2 & 0 \\
        \hline
        you kept everything inside and even though i tried it all fell apart what it meant to me will eventually be a memory of a time when i tried so hard & [Pre-Chorus] & Sad & 0 & -0.2 & -0.29 \\
        \hline
        i tried so hard and got so far but in the end it does not even matter i had to fall to lose it all but in the end it does not even matter & [Chorus] & N/A & 0 & 0 & -0.09 \\
        \hline
        i have put my trust in you pushed as far as i can go for all this there is only one thing you should know i have put my trust in you pushed as far as i can go for all this there is only one thing you & [Bridge]\_1 & N/A & 0 & 0 & 0.05 \\
        \hline
        should know & [Bridge]\_2 & N/A & 0 & 0 & 0 \\
        \hline
        i tried so hard and got so far but in the end it does not even matter i had to fall to lose it all but in the end it does not even matter & [Chorus] & N/A & 0 & 0 & -0.09 \\
        \hline
    \end{tabular}
    \end{adjustbox}
    \caption{Comprehensive view on sentiment and abuse detection for  Linkin Park's song "In The End"}
    \label{tab:lp_in_the_end_chunks}
\end{table*}





\end{document}